\pdfoutput=1
\documentclass[acmtog,nonacm]{acmart} 

\usepackage{pifont}
\newcommand{\cmark}{\ding{51}} 
\usepackage{kotex} 
\usepackage{booktabs} 
\usepackage{tabularx}
\usepackage{tabulary}
\usepackage{commath} 
\usepackage{mathtools} 

\usepackage{booktabs} 
\let\ul\underline 

\usepackage{wrapfig}
\usepackage{lipsum}
\usepackage[export]{adjustbox} 
\usepackage{subcaption} 
\usepackage{graphicx} 
\usepackage{harpoon} 
\usepackage{anyfontsize} 

\usepackage[ruled]{algorithm2e} 

\usepackage{xcolor} 
\definecolor{cadmiumgreen}{rgb}{0.0, 0.42, 0.24}
\definecolor{violet(ryb)}{rgb}{0.53, 0.0, 0.69}
\makeatletter
\DeclareRobustCommand{\rvs}{%
  \@bsphack
  \normalcolor
  \@esphack
}
\DeclareRobustCommand{\stoprvs}{%
  \@bsphack
  \normalcolor
  \@esphack
}
\makeatother

\SetAlFnt{\small}
\SetAlCapFnt{\small}
\SetAlCapNameFnt{\small}
\SetAlCapHSkip{0pt}

\setcopyright{acmcopyright}
\acmJournal{TOG}
\acmYear{2020}
\acmVolume{39}
\acmNumber{6}
\acmArticle{1}
\acmMonth{12}
\acmDOI{10.1145/3414685.3417797}

\begin{document}
\title{Neural Crossbreed: Neural Based Image Metamorphosis}

\author{Sanghun Park}
\affiliation{%
 \institution{KAIST, Visual Media Lab}
}
\email{reversi@kaist.ac.kr}

\author{Kwanggyoon Seo}
\affiliation{%
 \institution{KAIST, Visual Media Lab}
}
\email{skg1023@kaist.ac.kr}

\author{Junyong Noh}
\affiliation{%
 \institution{KAIST, Visual Media Lab}
}
\email{nohjunyong@kaist.ac.kr}


\renewcommand\shortauthors{S. Park et al}
\begin{abstract}
We propose Neural Crossbreed, a feed-forward neural network that can learn a semantic change of input images in a latent space to create the morphing effect. Because the network learns a semantic change, a sequence of meaningful intermediate images can be generated without requiring the user to specify explicit correspondences. In addition, the semantic change learning makes it possible to perform the morphing between the images that contain objects with significantly different poses or camera views. Furthermore, just as in conventional morphing techniques, our morphing network can handle shape and appearance transitions separately by disentangling the content and the style transfer for rich usability. We prepare a training dataset for morphing using a pre-trained BigGAN, which generates an intermediate image by interpolating two latent vectors at an intended morphing value. This is the first attempt to address image morphing using a pre-trained generative model in order to learn semantic transformation. The experiments show that Neural Crossbreed produces high quality morphed images, overcoming various limitations associated with conventional approaches. In addition, Neural Crossbreed can be further extended for diverse applications such as multi-image morphing, appearance transfer, and video frame interpolation.

\end{abstract} 
%
%

\ccsdesc[500]{Computing methodologies~Image processing}
\ccsdesc[300]{Computing methodologies~Artificial intelligence}

%

\keywords{Image morphing, neural network,
content and style disentanglement}

\begin{teaserfigure}
  \includegraphics[width=\textwidth]{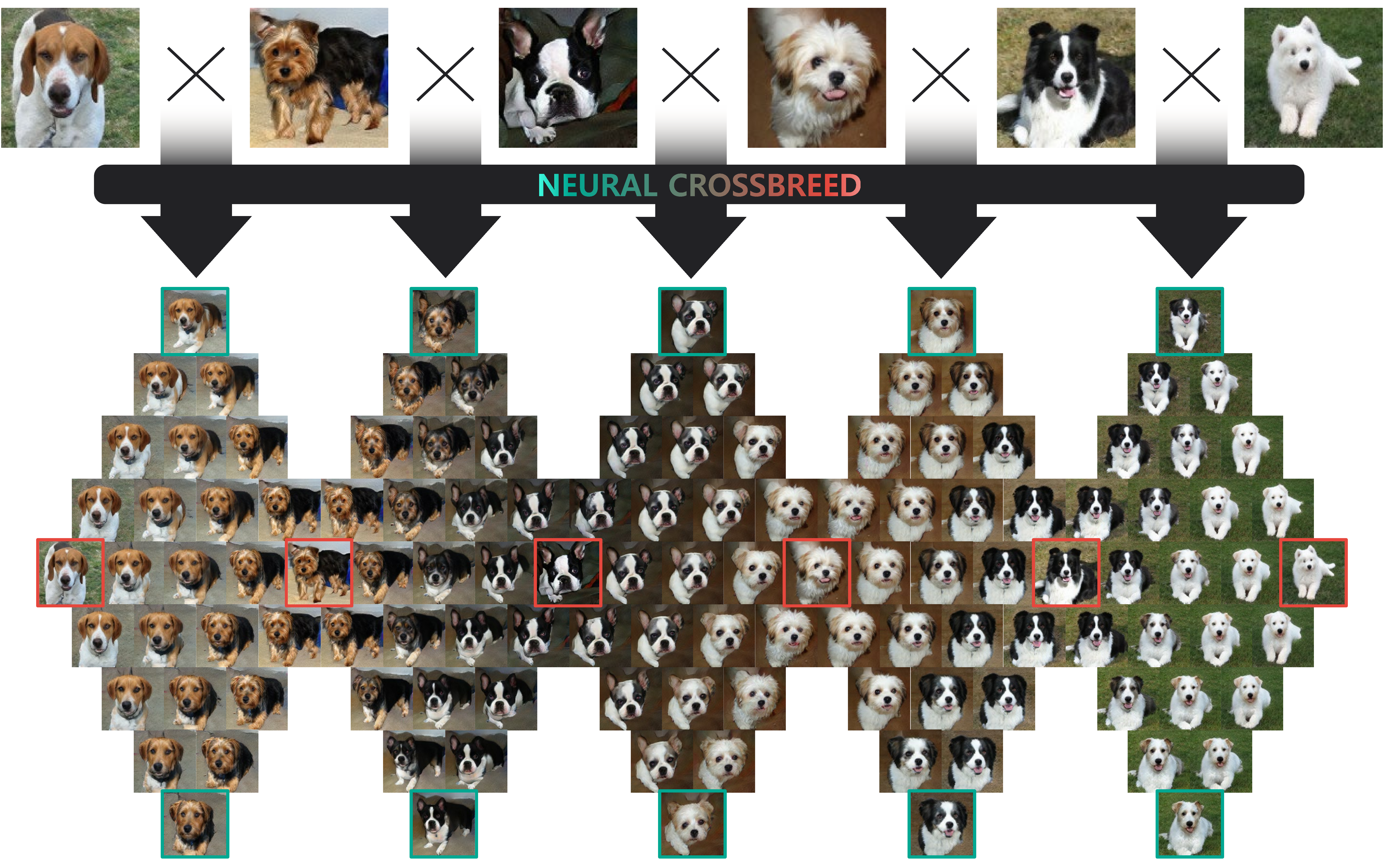}
  \caption{Neural Crossbreed creates a morphing effect given two input images (middle images between the red boxes). This  morphing transition can be extended to a 2D content-style transition manifold so that content and style
swapped images can also be generated (green boxes).}
  \Description{.}
  \label{fig:teaser}
\end{teaserfigure}

\maketitle


\section{Introduction}
To create a smooth transition between a given pair of images, conventional image morphing approaches typically go through the following three steps \cite{wolberg1998image}: establishing \textit{\textbf{correspondences}} between the two input images, applying a \textit{\textbf{warping}} function to make the shape changes of the objects in the images, and gradually \textit{\textbf{blending}} the images together for a smooth appearance transition. Because the warping and blending functions can be defined separately, they can be controlled individually with fine details \cite{liao2014automating,nguyen2015guiding,scherhag2019face}. 
Thanks to various interesting transformation effects it produces from one digital image to another, image morphing has been proven to be a powerful tool for visual effects in movies, television commercials, music videos, and games.

While conventional approaches to image morphing have long been studied and received much attention for the generation of interesting effects, they still suffer from some common limitations that prevent wider adoption of the technique. For example, conventional morphing algorithms often require the specification of \textit{\textbf{correspondences}} between the two images. In order to automate the process, most computer vision techniques assume that the two images are closely related and this assumption does not hold in many cases, so manual specification of the correspondences is usually inevitable. Another challenge is involved with occlusion of part of the object to which a \textit{\textbf{warping}} function is applied. A visible part of one image can be invisible in the other image and vice versa. This places a constraint in that the two images must have a perceptually similar semantic structure in terms of object pose and camera view when the user selects a pair of input images. Furthermore, two target objects with different textures are likely to produce an intermediate look of superimposed appearance due to a \textit{\textbf{blending}} operation even if their shapes are well aligned.

Recent research on learning-based image generative models has shown that high-quality images can be generated from random noise vectors \cite{kingma2013auto,goodfellow2014generative}. More recently, approaches to generating synthetic images that are indistinguishable from real ones \cite{brock2018large,karras2019analyzing} have also been reported. In this line of research, by walking in a latent space the model is capable of generating images with semantic changes \cite{radford2015unsupervised,jahanian2019steerability}. On the basis of this finding, we revisit the image morphing problem to overcome the limitations associated with conventional approaches by analyzing the semantic changes occurring by walking in a latent space. The manual specification of correspondences can be bypassed because semantic transformation can be learned in the latent manifold. Moreover, unlike traditional morphing techniques that sometimes distort the images to align the objects in them, the generative model creates proper intermediate images for the occluded regions in the reference images located at two different points in the latent space. In addition, unnatural blending of two textures can also be avoided.

Tackling the image morphing problems with the approach of walking in a latent space poses its own challenges. First, the generative model synthesizes images from randomly sampled high dimensional latent vectors; it is not clear how to associate user provided input images with the latent vectors that would generate the images. Second, it is extremely difficult to disentangle the vectors into each component that represents the shape and appearance separately, which is trivial in conventional morphing methods. Third, the generative model often produces low-fidelity images when the sampled latent vector is out of the range of the manifold.

In this paper, we propose Neural Crossbreed, a novel end-to-end neural network for image morphing, which overcomes the limitations of conventional morphing methods by learning the semantic changes in a latent space while walking. The contributions of our method are summarized as follows: 1) The morphing process is fully automatic and does not require any manual specification of correspondences. 2) The constraints on the object pose and camera view are much less restrictive compared to those required by conventional approaches when the user selects a pair of input images. 3) Although the morphing is performed in a latent space, the shape and appearance transitions can be handled separately for rich usability, which is similar to the benefits ensured by conventional approaches.
In addition, since our method uses a deep neural network, the inference time for the morphing between two images is much faster than that required by conventional methods that typically spend a long time to optimize the warping functions.
\begin{figure*}[ht]
  \includegraphics[width=\textwidth]{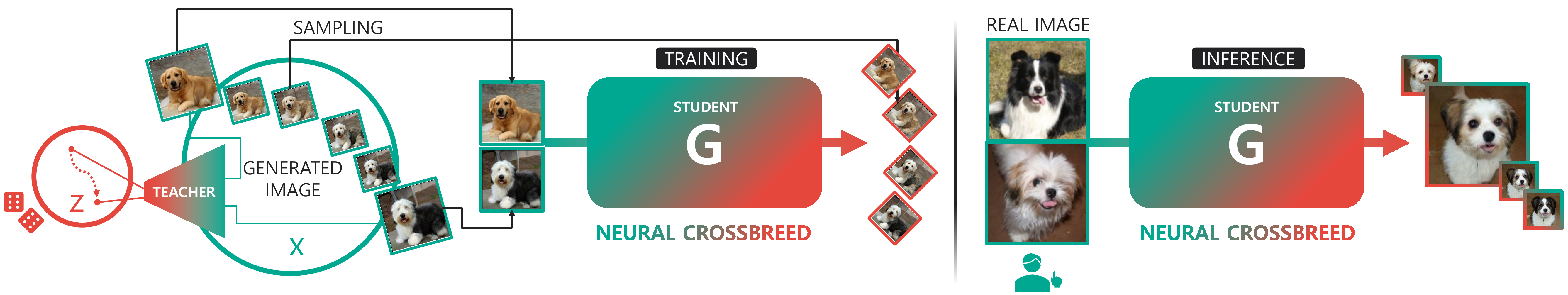}
  \caption{Overview of Neural Crossbreed. \rvs The training data is sampled from the pre-trained teacher generative model and used by the student network to learn the semantic changes of the generative model. Once training is complete, the student network can morph the real images entered by the user. \stoprvs}
  \label{fig:overview}
\end{figure*}

\section{Related Work}

\begin{table}[ht]
\caption{A comparison with previous approaches in terms of image morphing capability.}
\label{tab:difference}
\begin{tabulary}{\linewidth}{ L C C C }
\hline
                                       & Conventional image morphing & Image generation via latent space & Ours \\ \hline \hline
User input images                        & \cmark                 &                       & \cmark  \\ \hline
No specification of user correspondences &                        & \cmark                & \cmark  \\ \hline
Large occlusion handling                 &                        & \cmark                & \cmark  \\ \hline
Decoupled content and style transition   & \cmark                 &                       & \cmark  \\ \hline
\end{tabulary}
\end{table}

\textbf{Image Morphing }
Over the past three decades, image morphing has been developed in a direction that reduces user intervention in establishing correspondences between the two images \cite{wolberg1998image}. In the late 1980s, mesh warping that uses mesh nodes as pairs of correspondences was pioneered \cite{smythe1990two}. Thereafter, field morphing utilizing simpler line segments than meshes was developed \cite{beier1992feature}. A recent notable study performed the optimization of warping fields in a specific domain to reduce user intervention \cite{liao2014automating} but still required sparse user correspondences. Following this research direction, we propose an end-to-end network for automatic image morphing obviating the need for the user to specify corresponding primitives.

\textbf{Image Generation via Latent Space } 
Generative models have shown that semantically meaningful intermediate images can be produced by moving from one vector to another in a latent space \cite{radford2015unsupervised, brock2018large}. Recently, an interesting art tool called Ganbreeder \cite{ganbreeder} was introduced, which discovers new images by exploring a latent space randomly. Because Ganbreeder makes it necessary to associate the input image with a latent vector, it is difficult for the user to provide their own images as input, which limits practical use. To address this, a group of research focuses on embedding given images into the latent space of pre-trained generative models \cite{lipton2017precise,abdal2019image2stylegan,abdal2019image2stylegan++}. However, this approach often results in the loss of the details of the original image and requires a number of iterative optimizations that take a long time to embed a single image \cite{bau2019seeing}. In this paper, our network learns semantic image changes produced by random walking in a latent space while taking a form consisting of an encoder and decoder to accept user images. \rvs Table \ref{tab:difference} highlights the advantages of our method compared with previous approaches. \stoprvs

\textbf{Content and Style Disentanglement }
Traditional image morphing methods control the shape and appearance transitions individually by decoupling warping and blending operations \cite{liao2014automating,nguyen2015guiding,scherhag2019face}. A similar concept was introduced for the style transfer. For instance, it is possible to convert a photo to a famous style painting while preserving the content of the photo \cite{xie2007feature,gatys2015neural,huang2017arbitrary,kotovenko2019content}. Inspired by this concept, we designed a network that allows the user to perform decoupled \textit{content} and \textit{style} transitions; in this paper, we also use the same terminology of content and style transitions, as used in the work of \citet{xie2007feature}.

\textbf{Neural based Image Manipulation }
Some researchers tried to interpret the latent space of generative models to manipulate the image \cite{radford2015unsupervised,chen2016infogan,jahanian2019steerability,shen2019interpreting}. A challenge in this direction of research lies in how to associate user provided input images with latent vectors. One way to tackle this challenge is to manipulate images in a pre-trained feature space \cite{gardner2015deep,upchurch2017deep,lira2020ganhopper}. Meanwhile, image translation \cite{isola2017image} can also be considered as a style transfer that manipulates an image to adopt the visual style of another. Success of the style transfer often depends on how to translate the attributes of one image to those of another \cite{huang2018munit,lee2018drit,liu2019funit,lee2019dritpp,choi2020starganv2}. \rvs There are recent studies \cite{chen2019homomorphic,gong2019dlow,wu2019relgan} that allow changes in the characteristics of the object such as the expression or style in a single image. Despite the success of this translation, it is not clear how to smoothly change one image into another to achieve a morphing effect.

Concurrent to our work, \citet{viazovetskyi2020stylegan2} trained an image translation model that manipulates the image with synthetic data created by a generative network. Unfortunately, their method produces only a single transition result given two input images. Therefore, the model is not appropriate for image morphing. Another concurrent work by \citet{fish2020image} suggests a way to replace the warping and blending operation of the conventional morphing with a spatial transformer and an auto-encoder network, respectively. Similar to conventional methods, this method still suffers from ghosting artifacts when the content of the two images are very different. In contrast, our network progressively changes both content and style of one image into another without such artifacts.


\section{Morphing Dataset Preparation}

In the era of deep learning, using one pre-trained network to train another network has become popular. Examples include feature extraction \cite{johnson2016perceptual}, transfer learning \cite{yosinski2014transferable}, knowledge distillation \cite{hinton2015distilling}, and data generation \cite{karras2017progressive,such2019generative}. \rvs Concurrent to our work, \citet{viazovetskyi2020stylegan2} showed how to train a network for image manipulation using the data created by a pre-trained generative model. 
Similar to their strategy, we built a morphing dataset to distill \textit{\textbf{the ability for semantic transformation}} from a pre-trained ``teacher'' generative model to our ``student'' generator $G$.
\stoprvs
Figure \ref{fig:overview} illustrates the overview of Neural Crossbreed framework.

To build a morphing dataset, we adopt the BigGAN \cite{brock2018large} which is known to work well for class-conditional image synthesis. In addition to noise vectors sampled at two random points, the BigGAN takes a class embedding vector as input to generate an intended class image. This new image is generated by the linear interpolation of two noise vectors ${z_A, z_B}$ and two class embedding vectors ${e_A, e_B}$ according to control parameter $\alpha$. The interpolation produces a smooth semantic change from one class of image to another.

Sampling at two random points results in different image pairs in terms of appearance, pose, and camera view of an object. This effectively simulates a sufficient supply of a pair of user selected images under weak constraints on the posture of the object. As a result, we have class labels $\{l_{A},l_{B}\} \in L$ and training data triplets $\{x_{A},x_{B},x_{\alpha}\}$ for our generator to learn the morphing function. The triplets can be computed as follows:
\begin{align} \label{eq:data}
\begin{split}
x_A &= BigGAN(z_A, e_A) \text{, } x_B = BigGAN(z_B, e_B), \\
x_\alpha &= BigGAN \left( (1-\alpha)z_A + \alpha z_B, (1-\alpha)e_A + \alpha e_B) \right),
\end{split}
\end{align}
where $\alpha$ is sampled from the uniform distribution $\mathcal{U}$[0,1].

\section{Neural Based Image Morphing}
\subsection{Formulation for Basic Transition} \label{sec:basic} 
The aim is to train morphing generator $G:\{x_{A},x_{B},\alpha\} \rightarrow y_{\alpha}$ that produces a sequence of smoothly changing morphed images $y_{\alpha} \in Y$ from a pair of given images $\{x_{A},x_{B}\} \in X$, where $\alpha \{ \alpha \in \mathbb{R}: 0\leq\alpha\leq 1 \}$ is a control parameter. 
In the training phase, generator $G$ learns the ability to morph the images in a latent space of the teacher generative model. Note that a pair of input images $\{x_{A},x_{B}\}$ is sampled at two random points in the latent space, and a target intermediate image $x_{\alpha}$ is generated by linearly interpolating the two sampled points at an arbitrary parameter value $\alpha$. 
Generator $G$ learns the mapping from sampled input space $\{x_{A},x_{B},\alpha\}$ to target morphed image $x_{\alpha}$ of the teacher model by adversarially training discriminator $D$, where $D$ aims to distinguish between data distribution $x \in X$ and generator distribution $y \in Y$. After completion of the learning, generator $G$ becomes ready to produce a morphing output, which reflects the semantic changes of the teacher model, $x_{\alpha} \approx y_{\alpha}=G(x_{A},x_{B},\alpha)$. The user provides trained generator $G$ with two arbitrary images in the inference phase. Smooth transition between the two images is then possible by controlling parameter $\alpha$.

\subsection{Disentangled Content and Style Transition} \label{sec:disentangled}

Our basic generator $G$ makes a smooth transition for both the content and the style of image $x_{A}$ into those of $x_{B}$ as one combined process. In this section, we describe how basic generator $G$ can be extended to handle content and style transition separately for richer usability. For clarity, we first define the terms, \textit{content} and \textit{style} used in this paper. \textit{Content} refers to the pose, the camera view, and the position of the object in an image. \textit{Style} refers to the appearance that identifies the class or domain of the object. Now, we extend the notation described in Section \ref{sec:basic} to express the content and style of an image. Specifically, training data triplets $x_{A}$, $x_{B}$, and $x_{\alpha}$ are expanded to $x_{AA}$, $x_{BB}$, and $x_{\alpha\alpha}$ where the first subscript corresponds to content and the second subscript corresponds to style.


\begin{figure}[h!]
\begin{subfigure}{0.44\linewidth}
    \includegraphics[width=\linewidth]{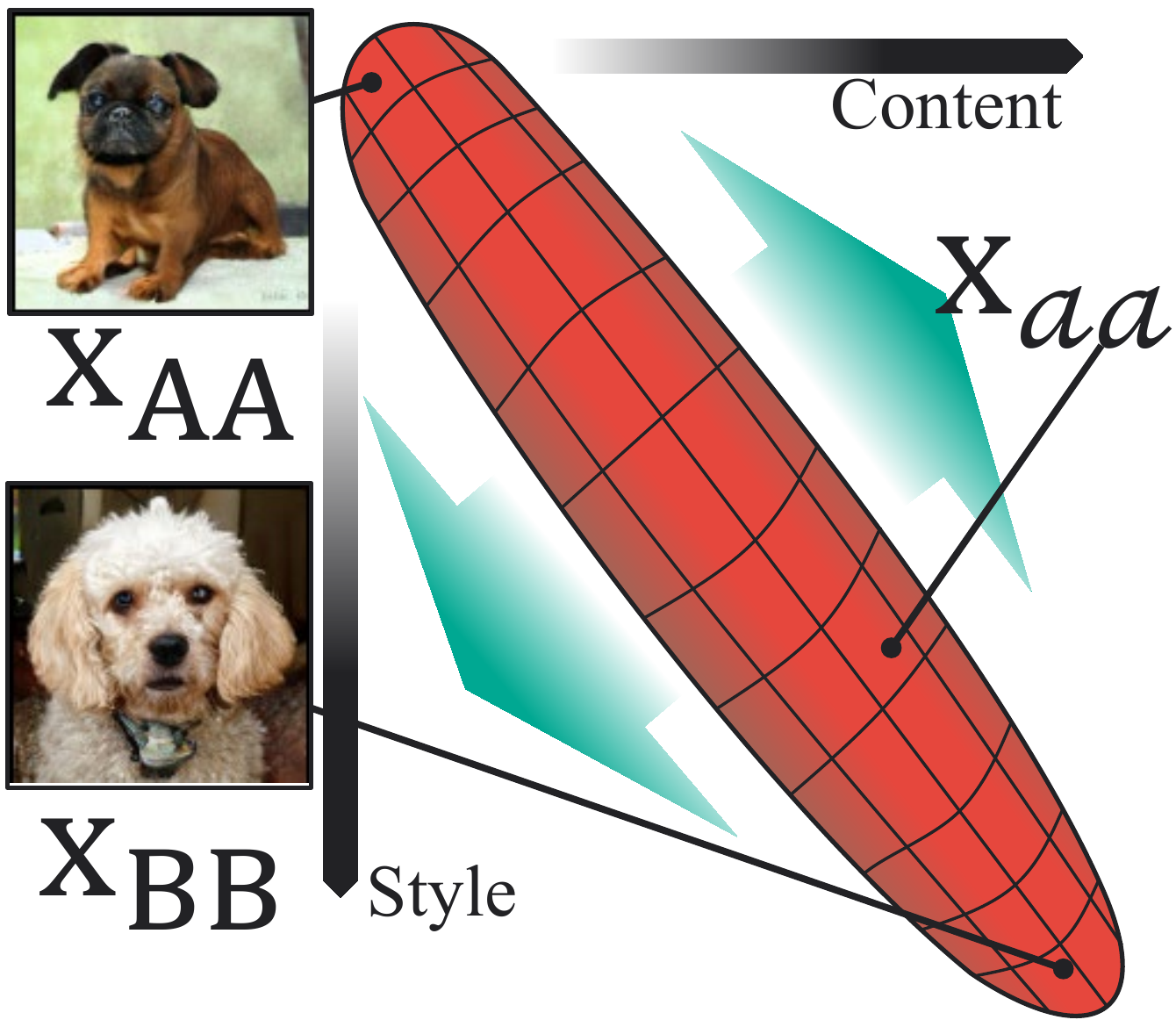} 
    \caption{Basic transition}
    \label{fig:disetangled_1}
\end{subfigure}
\begin{subfigure}{0.54\linewidth}
    \includegraphics[width=\linewidth]{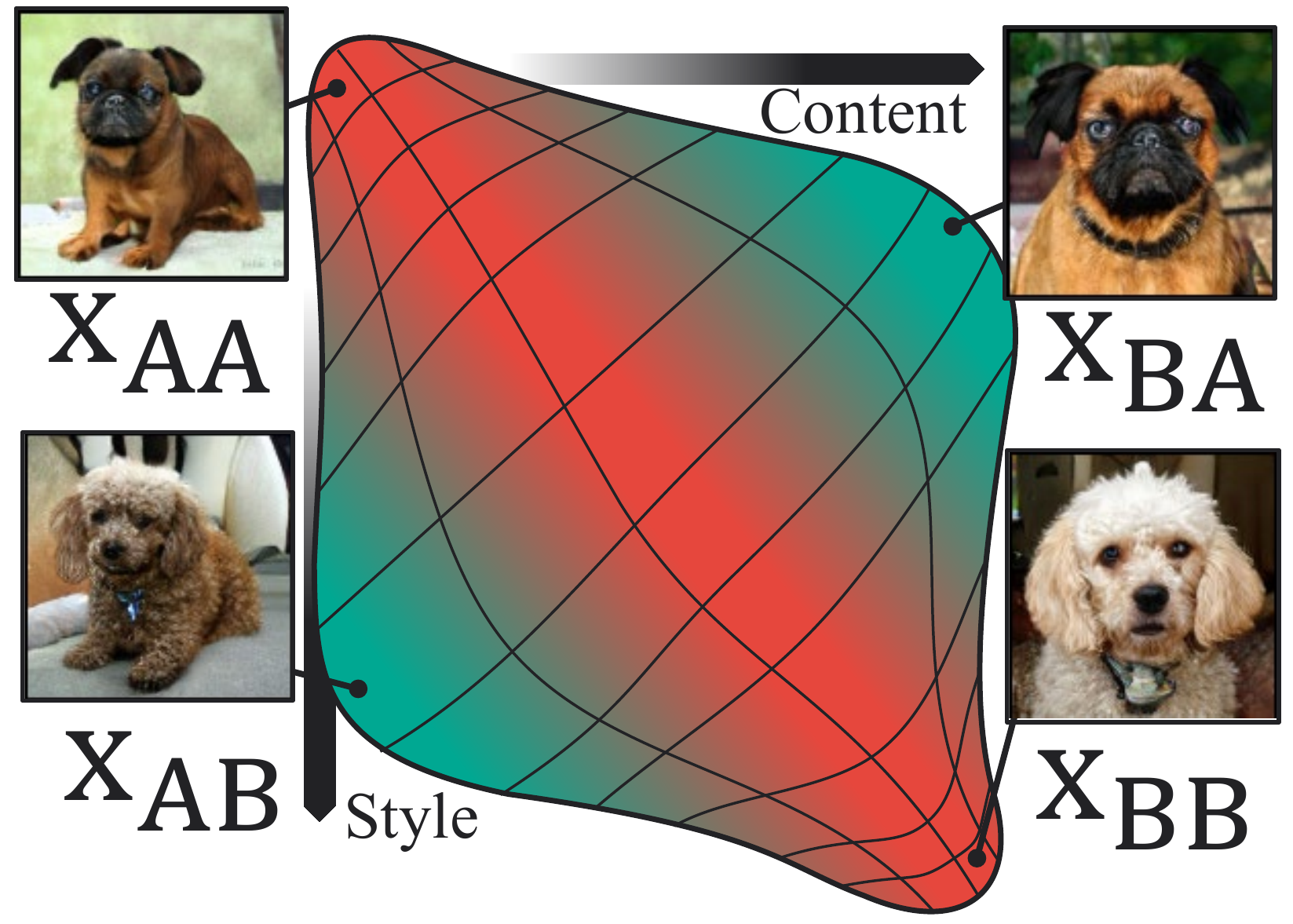}
    \caption{Disentangled transition}
    \label{fig:disetangled_2}
\end{subfigure}
\caption{Disentangled content and style transition. \rvs (a) Given a training data triplet $\{x_{AA},x_{BB},x_{\alpha\alpha}\}$, the generator learns a one-dimensional manifold that is capable of basic morphing by supervised and self-supervised learning. (b) The basic manifold is extended to a two-dimensional manifold through unsupervised learning of the swapped images of the content and style. \stoprvs}

\label{fig:content_style}
\end{figure}

The disentangled transition increases the dimension of control from a 1D transition manifold between the input images to a 2D transition manifold that spans the plane defined by content-style axes as shown in Figure \ref{fig:content_style}. This extension is achieved by training generator $G$ to additionally learn the transformation between the two images that have swapped style and content components $\{x_{AB}, x_{BA}\}$. Note, however, that our training data triplets $\{x_{AA},x_{BB},x_{\alpha\alpha}\}$ are only for basic transition, and do not contain any content and style of swapped images. Therefore, while the basic transition can be trained by the combination of a supervised and a self-supervised learning, the disentangled transitions can only be trained by unsupervised learning. Formally, we represent morphing generator $G$ as follows:
\begin{align} \label{eq:gen}
\begin{split}
y_{\alpha_{c}\alpha_{s}} &=  G(\rvs x_{AA},x_{BB}\stoprvs,\alpha_{c},\alpha_{s}) \quad\quad \\
\alpha_{c},\alpha_{s} &\in \mathbb{R}: 0\leq\alpha_{c},\alpha_{s}\leq 1 ,
\end{split}
\end{align}
where $\alpha_{c}$ and $\alpha_{s}$ indicate the control parameter for the content and style, respectively. In the inference phase, by controlling $\alpha_{c}$ and $\alpha_{s}$ individually, the user can smoothly change the content while fixing the style of the output image, and vice versa.

\subsection{Network Architecture}

\rvs
We based our network architecture on modern auto-encoders \cite{huang2018munit,lee2019dritpp,liu2019funit, choi2020starganv2} that are capable of learning content and style space. To achieve image morphing, we added a way to adjust latent codes according to morphing parameters. The latent code adjustment allows the student network to learn the basic morphing effects of the teacher network. Moreover, by individually manipulating the latent codes in a separate content and style space, the student network can learn disentangled morphing that the teacher network cannot create.
\stoprvs

\begin{figure}[h!]
  \includegraphics[width=\linewidth]{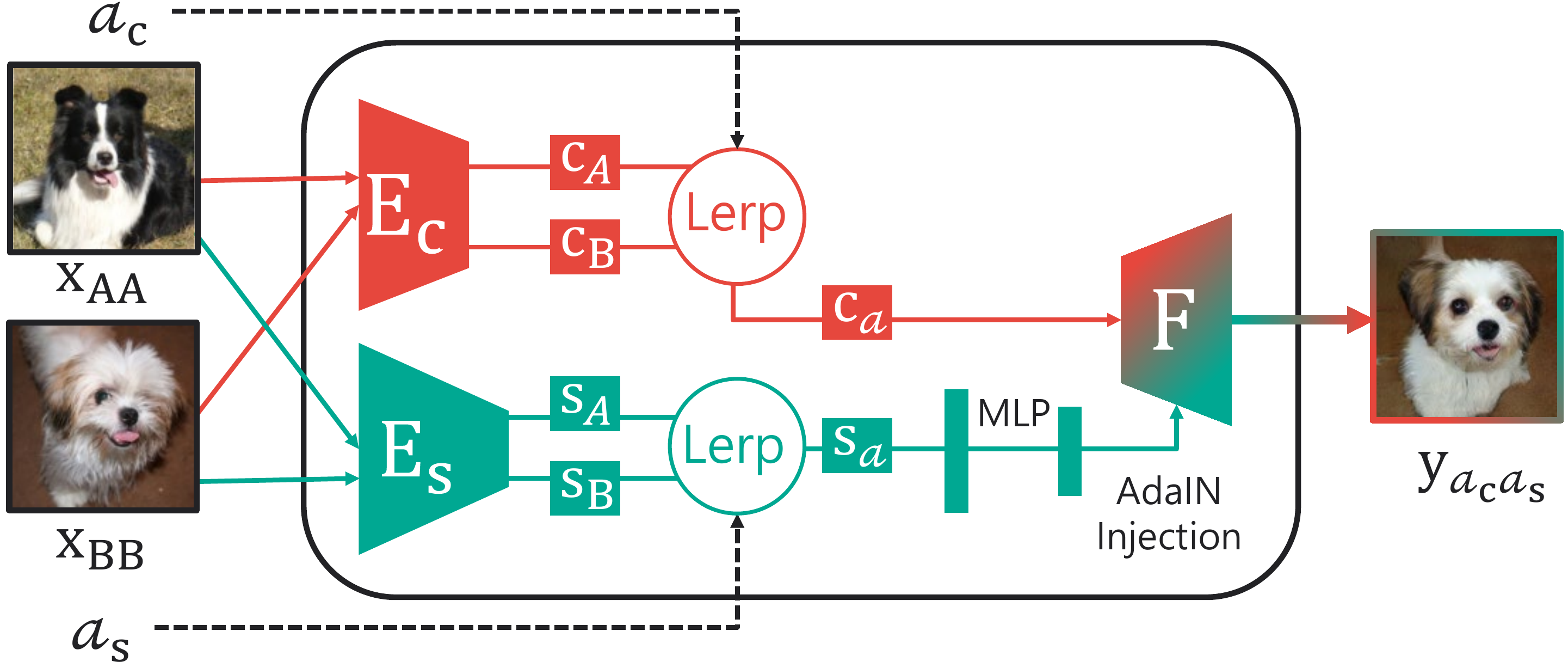}
  \caption{Network architecture. \rvs Our morphing generator encodes content and style codes $\{c_{A},c_{B},s_{A},s_{B}\}$ from two input images $\{x_{AA},x_{BB}\}$. The decoder combines the content and style codes $\{c_{\alpha_c},s_{\alpha_s}\}$ interpolated by the morphing parameter to produce morphed output $y_{\alpha_{c}\alpha_{s}}$. \stoprvs}
  \label{fig:network}
\end{figure}

We design a convolutional neural network $G$ that, given input images \rvs$\{x_{AA},x_{BB}\}$\stoprvs, outputs a morphed image $y_{\alpha_{c}\alpha_{s}}$, as illustrated in Figure \ref{fig:network}. Morphing generator $G$ consists of a content encoder $E_c$, a style encoder $E_s$, and a decoder $F$. $E_c$ extracts content code $c=E_c(x)$ and $E_s$ extracts style code $s=E_s(x)$ from input images $x$. The weights of each encoder are shared across different classes so that the input images can be mapped into a shared latent space associated with either content or style across all classes. $F$ decodes content code $c$ along with style code $s$ that has gone through a mapping network, which is a form of a multilayer perceptrons (MLP) \cite{karras2019style} followed by adaptive instance normalization (AdaIN) layers \cite{huang2017arbitrary} into output image $y$. Note that all of the components $E_c, E_s, F,$ and MLP of the generator learn the mappings across different classes using only a single network.

Equation \ref{eq:gen} can be factorized as follows:
\begin{align} \label{eq:factor1}
\begin{split}
y_{\alpha_{c}\alpha_{s}} &= G(x_{AA}, x_{BB},\alpha_{c},\alpha_{s}) \\
                         &= F(c_{\alpha_c}, s_{\alpha_s})\footnotemark, \\
\end{split}
\end{align}
\begin{align} \label{eq:factor2}
\begin{split}
\text{where }
c_{\alpha_c} = Lerp(c_A, c_B, \alpha_c) &\text{, } s_{\alpha_s} = Lerp(s_A, s_B, \alpha_s), \\
        c_A = E_c(x_{AA}) &\text{, } c_B = E_c(x_{BB}), \\
        s_A = E_s(x_{AA}) &\text{, } s_B = E_s(x_{BB}).
\end{split}
\end{align}
\footnotetext{The mapping network and the AdaIN injection are omitted in all of the equations for simplicity.}
Here, $Lerp(A, B, \alpha)$ is a linear function that can be expressed as $(1-\alpha) A + \alpha B$ , whose role is to interpolate the latent codes extracted from each of the two encoders before feeding them into decoder $F$. The adoption of the $Lerp(\cdot)$ operator in our network is inspired by the hypothesis presented in \citet{bengio2013better}, which claims  ``deeper representations correspond to more unfolded manifolds and interpolating between points on a flat manifold should stay on the manifold''. By interpolating the content and style features in the deep bottleneck position of the network layers, generator network $G$ learns a mapping that produces a morphed image on the same manifold defined by the two input images. 

$D$ is a multi-task discriminator \cite{mescheder2018training,liu2019funit}, which produces $\abs{L}$ outputs for multiple adversarial classification tasks. The output of discriminator $D$ contains a real/fake score for each class. When training, only the $l$-th output branch of $D$ is penalized according to class labels $\{l_{A},l_{B}\} \in L$ associated with input images $\{x_A,x_B\}$. Through the adversarial training, discriminator $D$ enforces generator $G$ to extract the content and the style of the input images, and to smoothly translate one image into another. 
\rvs We provide architectural details in Appendix \ref{appx:architecture}. \stoprvs

\subsection{Training Objectives} \label{sec:objectives}
\textbf{Loss Functions }
We use two primary loss functions to train our network. The first function is an adversarial loss. Given a class label $l \in L$, discriminator $D$ distinguishes whether the input image is real data $x \in X$ of the class or generated data $y \in Y$ from $G$. On the other hand, generator $G$ tries to deceive discriminator $D$ by creating an output image $y$ that looks similar to a real image $x$ via a conditional adversarial loss function
\begin{align} \label{eq:adv}
L_{adv}(x,y,l) &= \mathbb{E}_{x}\big[\log D(x|l)\big] + \mathbb{E}_{y}\big[\log (1-D(y|l))\big].
\end{align}

The second function is a pixel-wise reconstruction loss. Given training data triplets $x$, generator $G$ is trained to produce output image $y$ that is close to the ground truth in terms of a pixel-wise L1 loss function. For brevity of the description, the pixel-wise reconstruction loss is simplified as follows:
\begin{align} \label{eq:pix}
\norm{x-y}_1 &\triangleq \mathbb{E}_{x,y}\Big[\norm{x-y}_1\Big].
\end{align}

\textbf{Identity Loss } 
Image morphing is a linear transformation between two input images $\{x_{AA}, x_{BB}\}$. To achieve this, generator $G$ should be able to convert the two input images as well as intermediate images into a representation defined on a deep unfolded manifold, and restore them back to the image domain. Generator $G$ learns to produce the images identical to the input in a self-supervised manner as follows:
\begin{align} \label{eq:idt}
\begin{split}
L^{idt}_{adv} &= L_{adv}(x_{AA}, y_{AA}, l_A) + L_{adv}(x_{BB}, y_{BB}, l_B), \\
L^{idt}_{pix} &= \norm{x_{AA}-y_{AA}}_1 + \norm{x_{BB}-y_{BB}}_1, \\
\text{where }
y_{AA} &= F(E_c(x_{AA}), E_s(x_{AA})) = G(x_{AA},x_{BB},0,0), \\
y_{BB} &= F(E_c(x_{BB}), E_s(x_{BB})) = G(x_{AA},x_{BB},1,1), \\
\end{split}
\end{align}
where identity images $\{y_{AA},y_{BB}\}$ are the restored back version of input images $\{x_{AA}, x_{BB}\}$. These identity losses help two encoders $E_c$ and $E_s$ extract latent codes onto a flat manifold from input image $x$, and also help decoder $F$ restore these two codes back in the form of input images.

\textbf{Morphing Loss } 
A result morphed from two images should have the characteristics of both of the images. To achieve this, we design a weighted adversarial loss that makes the output image have a weighted style between $x_{AA}$ and $x_{BB}$ according to the value of $\alpha$. In addition, given target morphing data $x_{\alpha\alpha}$, generator $G$ should learn the semantic change between the two input images by minimizing a pixel-wise reconstruction loss. The two losses can be described as follows:
\begin{align} \label{eq:morph}
\begin{split}
L^{mrp}_{adv} &= (1-\alpha) L_{adv}(x_{AA}, y_{\alpha\alpha}, l_A) + \alpha L_{adv}(x_{BB}, y_{\alpha\alpha}, l_B), \\
L^{mrp}_{pix} &= \norm{x_{\alpha\alpha}-y_{\alpha\alpha}}_1, \\
\text{where }
y_{\alpha \alpha} &= F(c_\alpha, s_\alpha) = G(x_{AA},x_{BB},\alpha,\alpha),
\end{split}
\end{align}
where $c_\alpha$ and $s_\alpha$ are interpolated from the extracted latent codes using parameter $\alpha$ and Equation \ref{eq:factor2}. Similar to the identity losses, these morphing losses enforce output images $y_{\alpha\alpha}$ from the generator to be placed on a line manifold connecting the two input images.

\textbf{Swapping Loss }
In order to extend the basic transition to a disentangled transition, we train generator $G$ to output content and style swapped images $\{y_{AB},y_{BA}\}$. However, since our training dataset does not include such images, the swapping loss should be minimized in an unsupervised manner as follows:
\begin{align} \label{eq:swap}
\begin{split}
L^{swp}_{adv} &= L_{adv}(x_{AA}, y_{BA}, l_A) + L_{adv}(x_{BB}, y_{AB}, l_B), \\
\text{where }
y_{AB} &= F(E_c(x_{AA}), E_s(x_{BB})) = G(x_{AA},x_{BB},0,1), \\
y_{BA} &= F(E_c(x_{BB}), E_s(x_{AA})) = G(x_{AA},x_{BB},1,0).
\end{split}
\end{align}
This loss encourages two encoders $E_c, E_s$ to separate the two characteristics from each input image in which content and style are entangled.

\textbf{Cycle-swapping Loss }
To further enable generator G to learn a disentangled representation and also guarantee that the swapped images preserve the content of the input images, we employ a cycle consistency loss \cite{zhu2017cyclegan,lee2019dritpp} as follows:
\begin{align} \label{eq:cyc}
\begin{split}
L^{cyc}_{adv} &= L_{adv}(x_{AA}, y'_{AA}, l_A) + L_{adv}(x_{BB}, y'_{BB}, l_B), \\
L^{cyc}_{pix} &= \norm{x_{AA}-y'_{AA}}_1 + \norm{x_{BB}-y'_{BB}}_1, \\
\text{where }
y'_{AA} &= F(E_c(y_{AB}), E_s(x_{AA})) = G(y_{AB},x_{AA},0,1), \\
y'_{BB} &= F(E_c(y_{BA}), E_s(x_{BB})) = G(y_{BA},x_{BB},0,1).
\end{split}
\end{align}
Here, cycle-swapped images $\{y'_{AA},y'_{BB}\}$ are the restored back version of input images $\{x_{AA}, x_{BB}\}$ through reswapping the content and style of swapped images $\{y_{AB}, y_{BA}\}$.

\textbf{Full Objective }
The full objective function is a combination of the two main terms. We train generator $G$ by solving the minimax optimization given as follows:
\begin{align} \label{eq:full}
\begin{split}
G^* &= \arg \min_G \max_D  \mathcal{L}_{adv} + \lambda \mathcal{L}_{pix} , \\
\text{where }
\mathcal{L}_{adv} &= L^{idt}_{adv} + L^{mrp}_{adv} + L^{swp}_{adv} + L^{cyc}_{adv}, \\
\mathcal{L}_{pix} &= L^{idt}_{pix} + L^{mrp}_{pix} + L^{cyc}_{pix}.
\end{split}
\end{align}
where the role of hyperparameter $\lambda$ is to balance the importance between the two primary loss terms.

In summary, the identity and the morphing losses are introduced first for the basic transition described in Section \ref{sec:basic}, while the swapping and the cycle-swapping losses are added for the disentangled transition described in Section \ref{sec:disentangled}. Note that during training, morphing parameters $\alpha_c,\alpha_s$ for generator $G$ are set to the same value in Equations \ref{eq:idt} and \ref{eq:morph} (i.e. $\alpha_c = \alpha_s = 0, \alpha, \text{ or } 1$ ) to learn the basic transition, but they are set differently in Equations \ref{eq:swap} and \ref{eq:cyc} to learn the disentangled transition. At the end of the training, the user can create a basic morphing transition by setting two parameters $\alpha_c,\alpha_s$ to the same value between 0 and 1 in the inference phase. The values of $\alpha_{c}$ and $\alpha_{s}$ can also be set individually for separate control of content and style because $G$ also learned the 2D content-style manifold.
\section{Experiments}

\subsection{Experimental Details}

\textbf{Training Data}
We created four types of morphing datasets using the images synthesized by BigGAN for experiments: \textsc{Dogs} (118 dog classes), \textsc{Birds} (39 aves classes), \textsc{Foods} (19 food classes), and \textsc{Landscapes} (8 landform classes). Each dataset corresponds to a subcategory of ImageNet \cite{ILSVRC15}. Note that all of the data were generated from the pre-trained BigGAN network at runtime in the training phase. The truncation of BigGAN was empirically set to 0.25 to generate training images with high fidelity and to allow the range of sample images wide enough for the effective learning by the generator. \rvs Examples of the training data generated by the teacher generative model are given in the first row of Figure \ref{fig:comparison_teacher}. \stoprvs

\textbf{Hyperparameters and Training Details }
All of our results were produced using the same default parameters for the training. We set hyperparameter $\lambda=0.1$ and used Adam optimizer \cite{kingma2014adam} with $beta_{1}=0.5$ and $beta_{2}=0.999$. To improve the rate of convergence, we employed Two Timescale Update Rule (TTUR) \cite{heusel2017gans} with learning rates of $0.0001$ and $0.0004$ for the generator and the discriminator, respectively. We used the hinge version of the adversarial loss \cite{lim2017geometric,tran2017hierarchical} with the gradient penalty on real data \cite{mescheder2018training}. The batch size was set to 16 and the model was trained for 200k iterations which required approximately four days of computation on four GeForce GTX 1080 Ti GPUs.

\textbf{Truncation Trick}
We remapped $\alpha_c$ and $\alpha_s$ using a truncation trick \cite{marchesi2017megapixel,pieters2018comparing,kingma2018glow,brock2018large,karras2019style} to confine the output space to a reasonable range at the inference stage. This trick is known to control the trade-off between image diversity and visual fidelity. 
\rvs Appendix \ref{appx:truncation} provides detailed explanations on how the truncation trick can be performed with a parameter $\tau$. \stoprvs
Note that the truncation is applied only to disentangled results described in this section. In addition, $\tau$ is specified in the caption of all of the disentangled results if the truncation trick was applied.


\textbf{Evaluation Protocol}
We used two datasets to explore effectiveness of the morphing network: ImageNet \cite{ILSVRC15} and AFHQ \cite{choi2020starganv2}.
At the inference stage, we sampled images from the ImageNet dataset for visual evaluation.
An ablation test and quantitative comparisons were performed using the dog images from AFHQ. Since AFHQ does not contain a label for each image, we constructed a pseudo label using a pre-trained classification model  \cite{wslimageseccv2018} trained on the Imagenet with the ResNeXt-101 32x16d backbone \cite{Xie2016}. We excluded the data outside \textsc{Dogs} categories and further removed the data with low confidence.
To measure the quality of the morphed image,
\rvs we employed Fr\`{e}chet Inception Distance (FID) \cite{heusel2017gans} as a performance metric and compared the statistics of the generated samples to those of real samples in AFHQ.
We set the value of $\alpha=0.5$ for the morphed result given two input images.
In addition, we measured the quality of the reconstructed image when $\alpha=0$ and $\alpha=1$ using the input image sampled from AFHQ as the ground truth. We utilized full-reference quality metrics, namely, Mean Squared Error (MSE), Peak Signal-to-Noise Ratio (PSNR), and Structural Similarity Index (SSIM), and a deep learning based metric, namely, Learned Perceptual Image Patch Similarity (LPIPS) \cite{zhang2018perceptual} \stoprvs


\subsection{Basic Morphing}
\begin{figure*}[!h]
  \includegraphics[width=\linewidth]{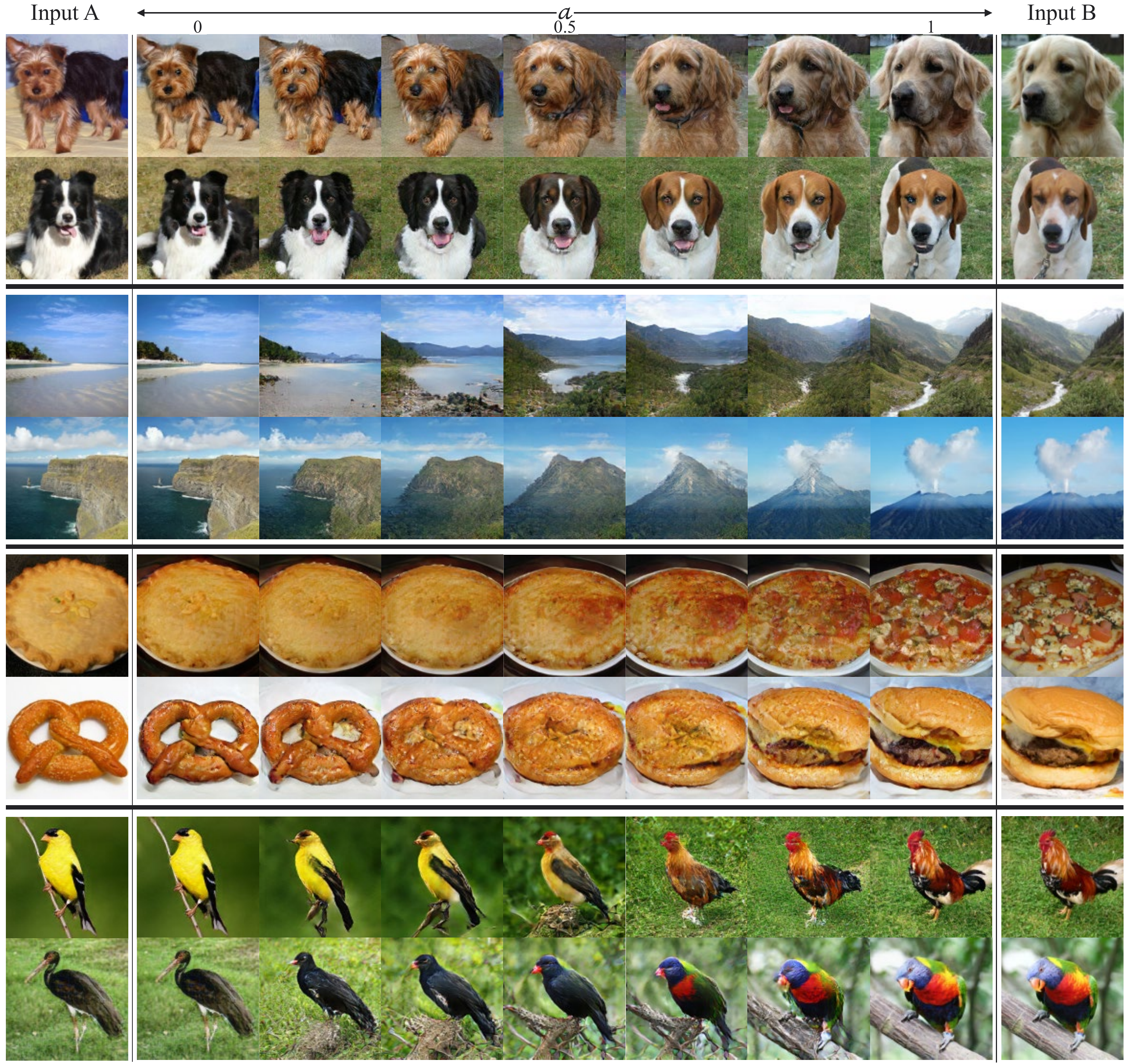}
  \caption{Results from the basic image morphing. \rvs Each block of two rows shows an example of morphing from the generator trained with \textsc{Dogs}, \textsc{Landscapes}, \textsc{Foods}, and \textsc{Birds} data. \stoprvs}
  \label{fig:morphing}
\end{figure*}

Figure \ref{fig:morphing} shows several basic morphing results from Neural Crossbreed trained on the dataset of \textsc{Dogs}, \textsc{Landscapes}, \textsc{Foods}, and \textsc{Birds}.
These results clearly verify that Neural Crossbreed can generate smoothly changing intermediate images according to the variation of morphing parameter $\alpha$. For example, the top row shows that the entire body shot of the silky terrier is transitioned to the head shot of the golden retriever in a semantically correct way albeit no correspondences were specified. The third row shows disappearance of the beach while a stream is naturally forming in the valley as the morphing progresses. The last row shows a set intermediate birds with features from both of the input images. Their shapes and appearances are very realistic with reasonable poses and vivid colors.

\subsection{Content and Style Disentangled Morphing} \label{sec:disen_result}
\begin{figure}[!h]
    \includegraphics[width=\linewidth]{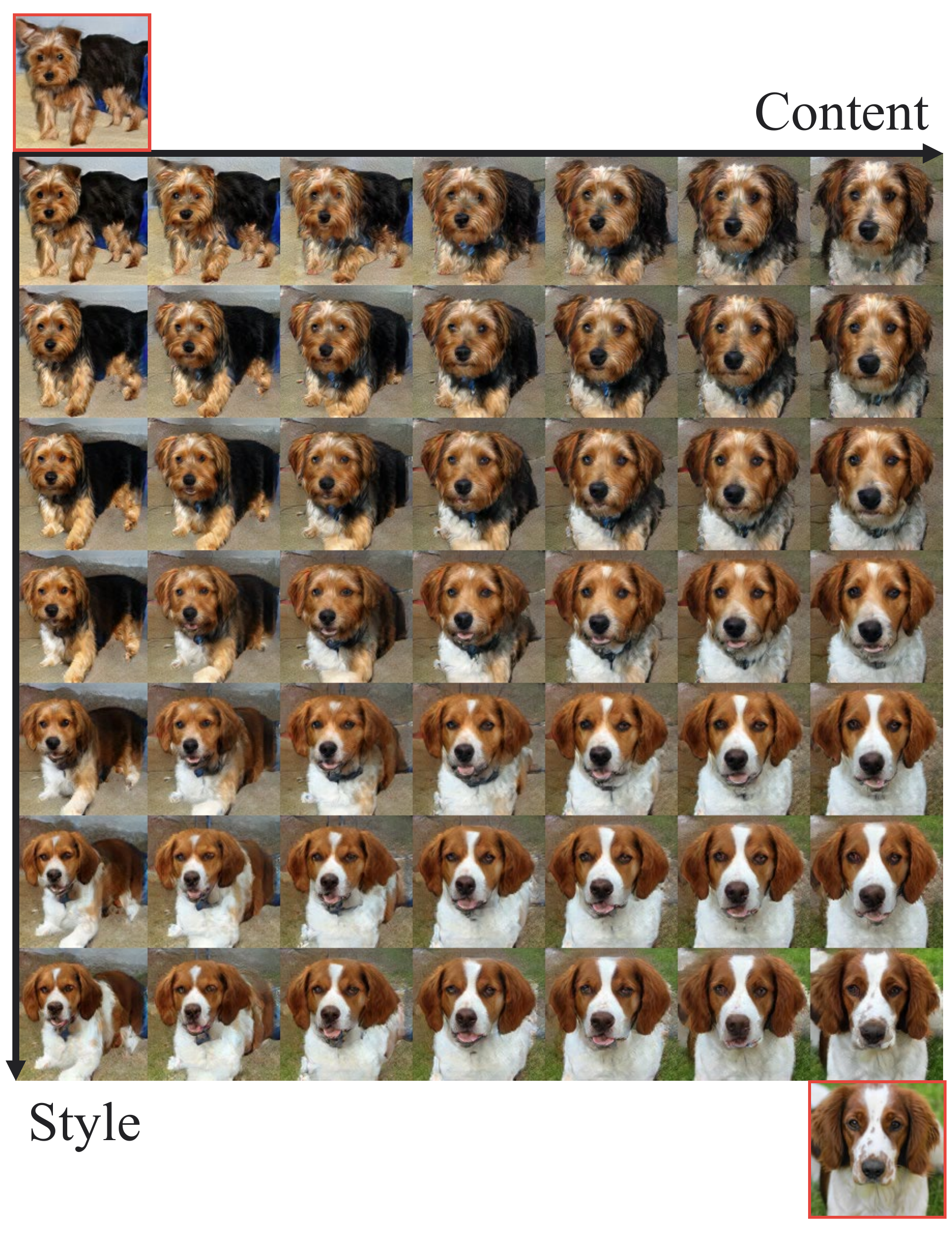}
    \caption{Visualization of the 2D content-style manifold created by two input images. The red boxes indicate the input images. Here, we use the truncation trick with $\tau = 0.3$. 
    \rvs Results produced without using the truncation trick can be found in Appendix \ref{appx:truncation}. \stoprvs
    }
\label{fig:disentangled}
\end{figure}
Warping and blending based approaches can make a decoupled shape and appearance transition. Neural Crossbreed can similarly control the content and style transition by individually adjusting $\alpha_c$ and $\alpha_s$. Figure \ref{fig:disentangled} visualizes the 2D transition manifold that spans the plane defined by the style and content of the two input images. Each row shows a transition from the entire body shot to the chest shot of a dog along the content axis. Note that the change in the diagonal direction corresponds to the basic transition. On the other hand, each column shows a transition from a silky terrier to a welsh springer spaniel along the style axis. 

\begin{figure*}[!h]
\begin{subfigure}{0.24\linewidth}
    \includegraphics[width=\linewidth]{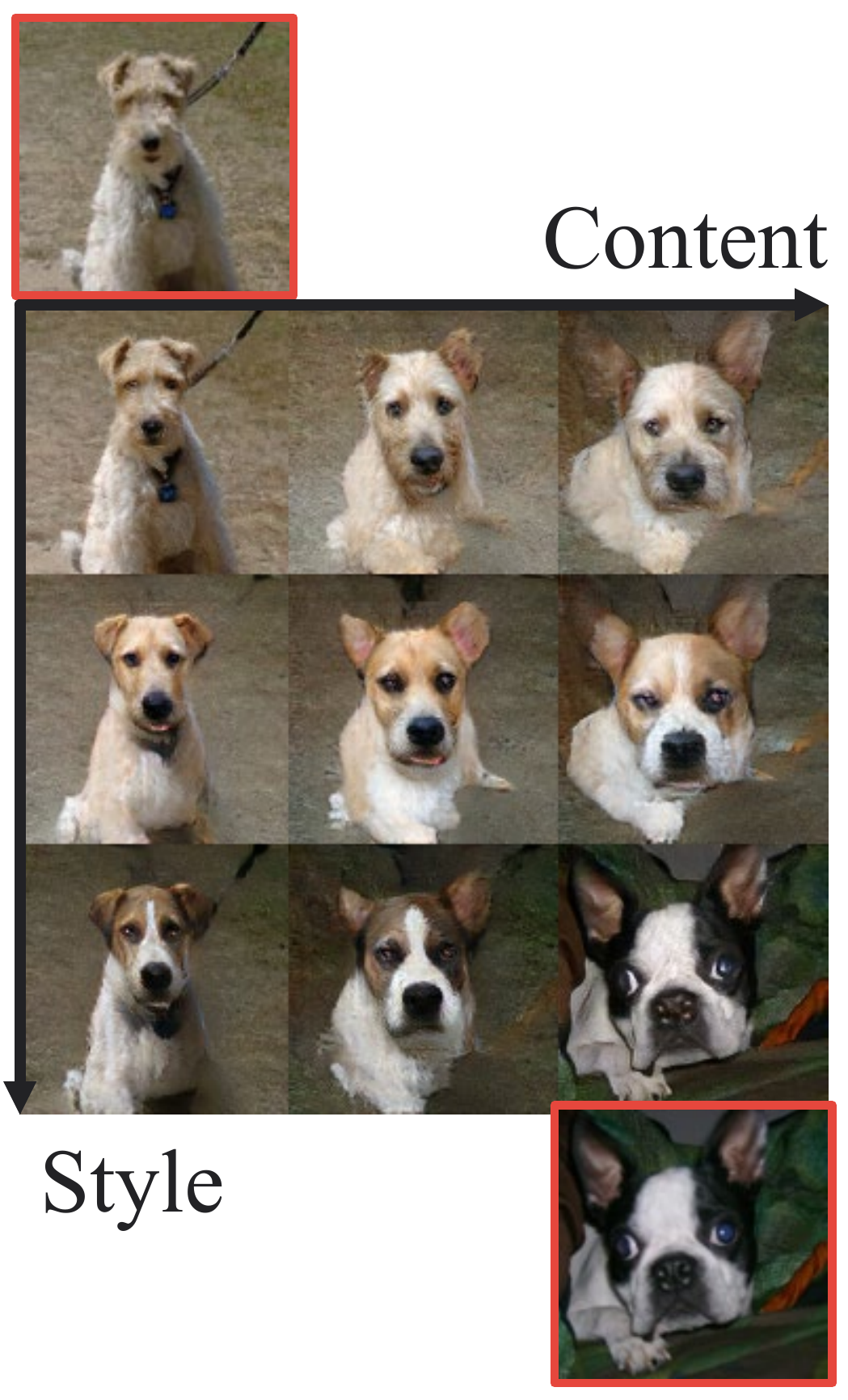} 
    \caption{w/o Identity loss}
    \label{fig:abl_no_idt}
\end{subfigure}
\begin{subfigure}{0.24\linewidth}
    \includegraphics[width=\linewidth]{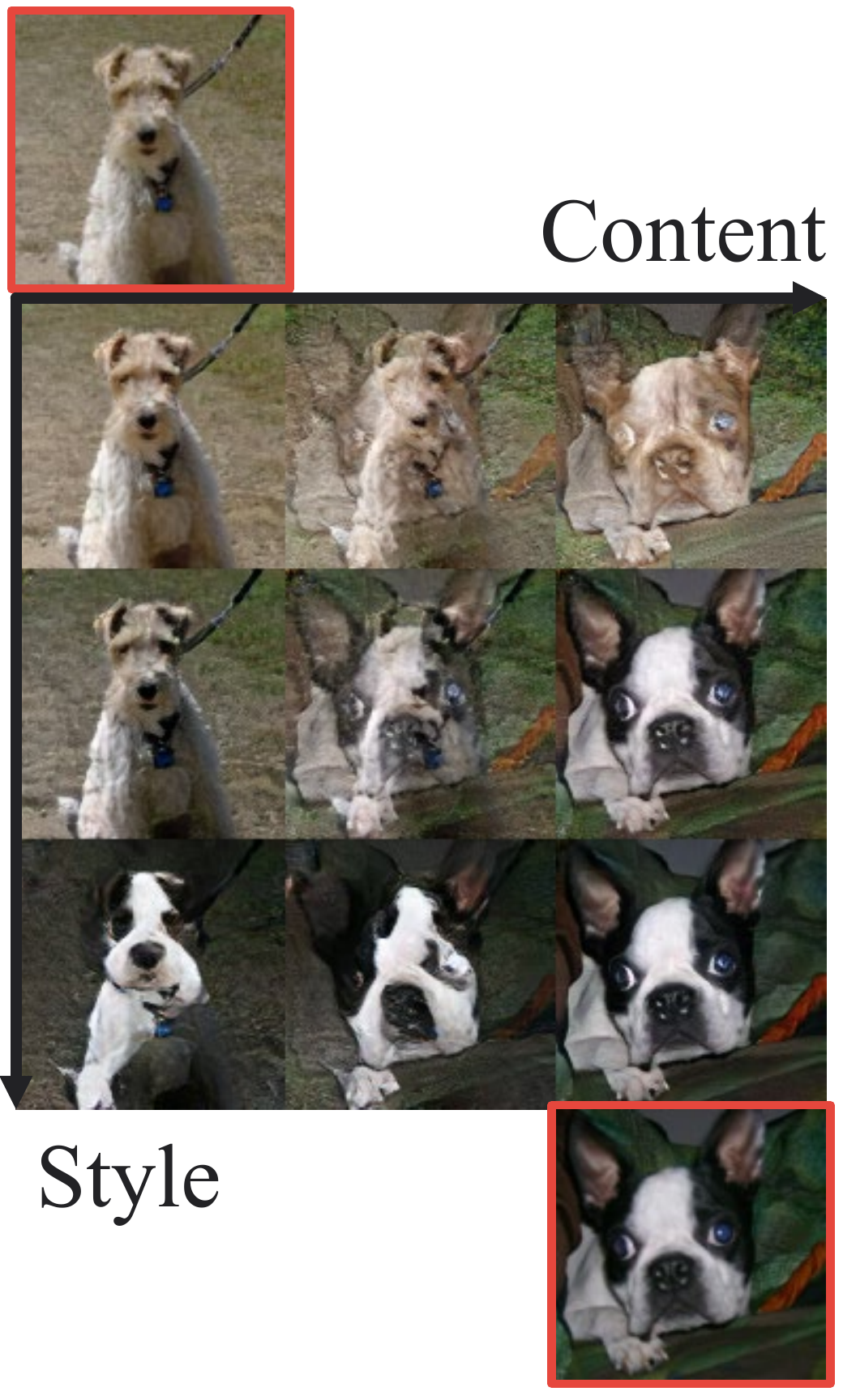}
    \caption{w/o morphing loss}
    \label{fig:abl_no_mrp}
\end{subfigure}
\begin{subfigure}{0.24\linewidth}
    \includegraphics[width=\linewidth]{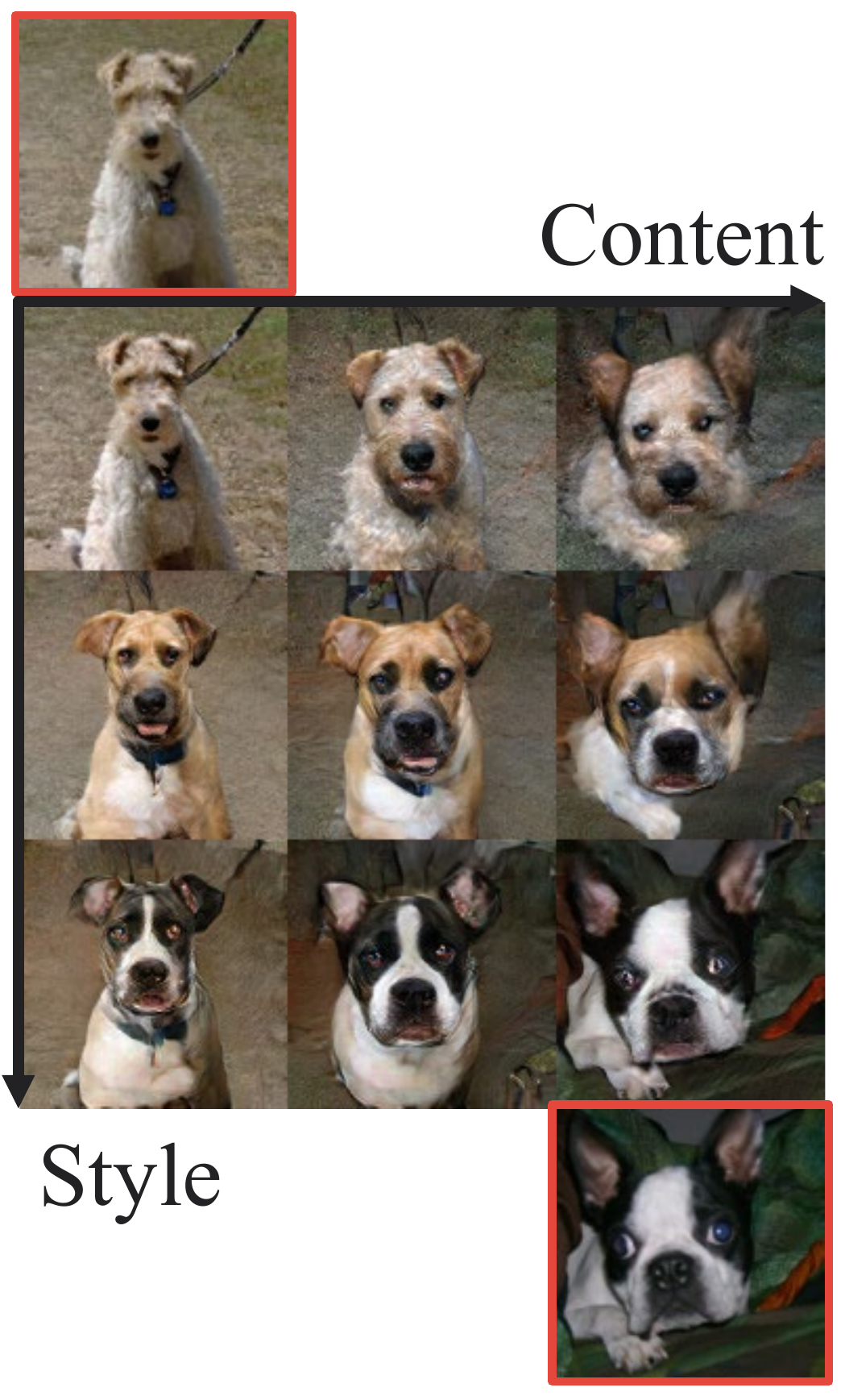}
    \caption{w/o swapping loss}
    \label{fig:abl_no_swp}
\end{subfigure}
\begin{subfigure}{0.24\linewidth}
    \includegraphics[width=\linewidth]{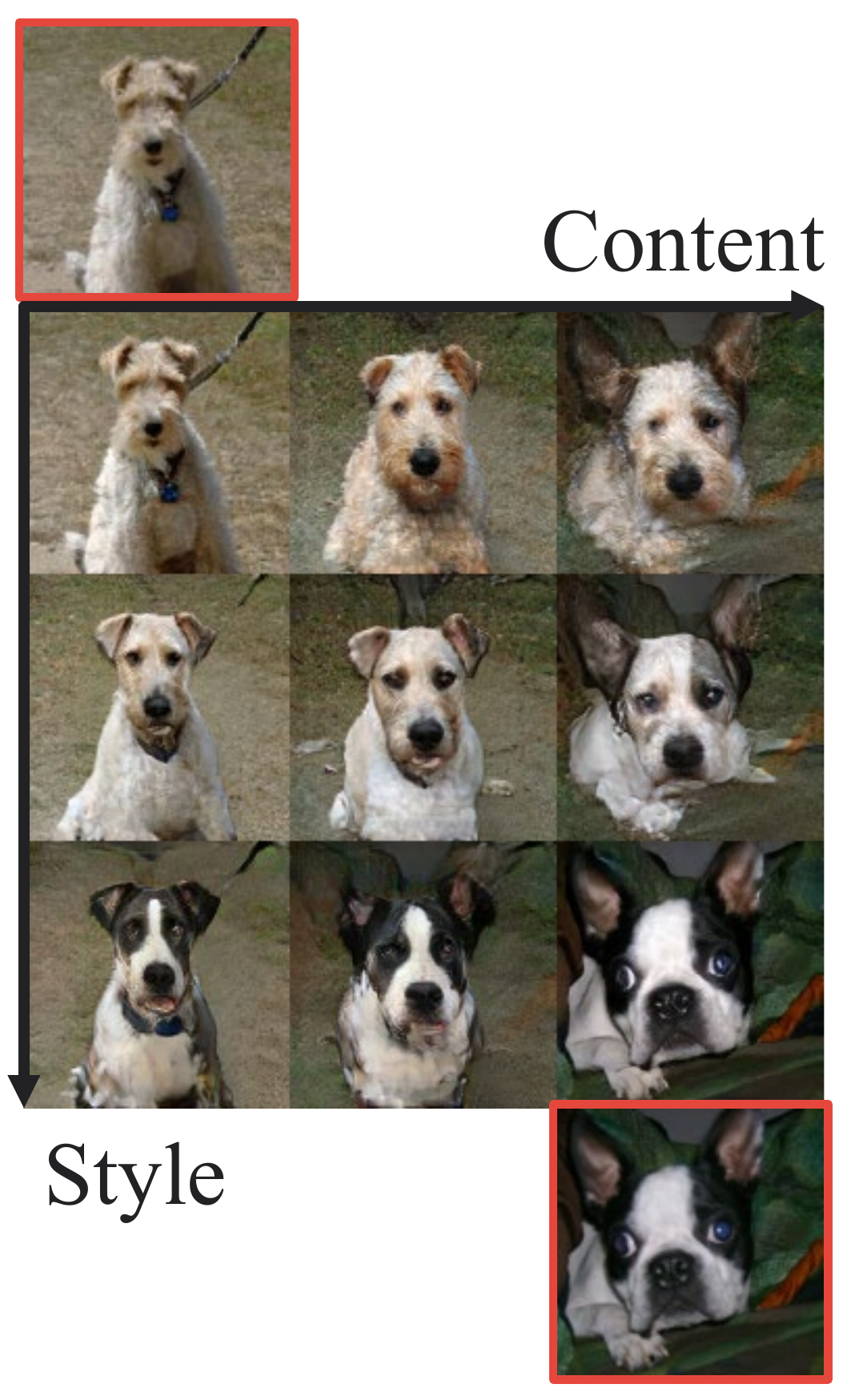}
    \caption{w/o cycle-swapping loss}
    \label{fig:abl_no_cyc}
\end{subfigure}
\begin{subfigure}{0.24\linewidth}
    \includegraphics[width=\linewidth]{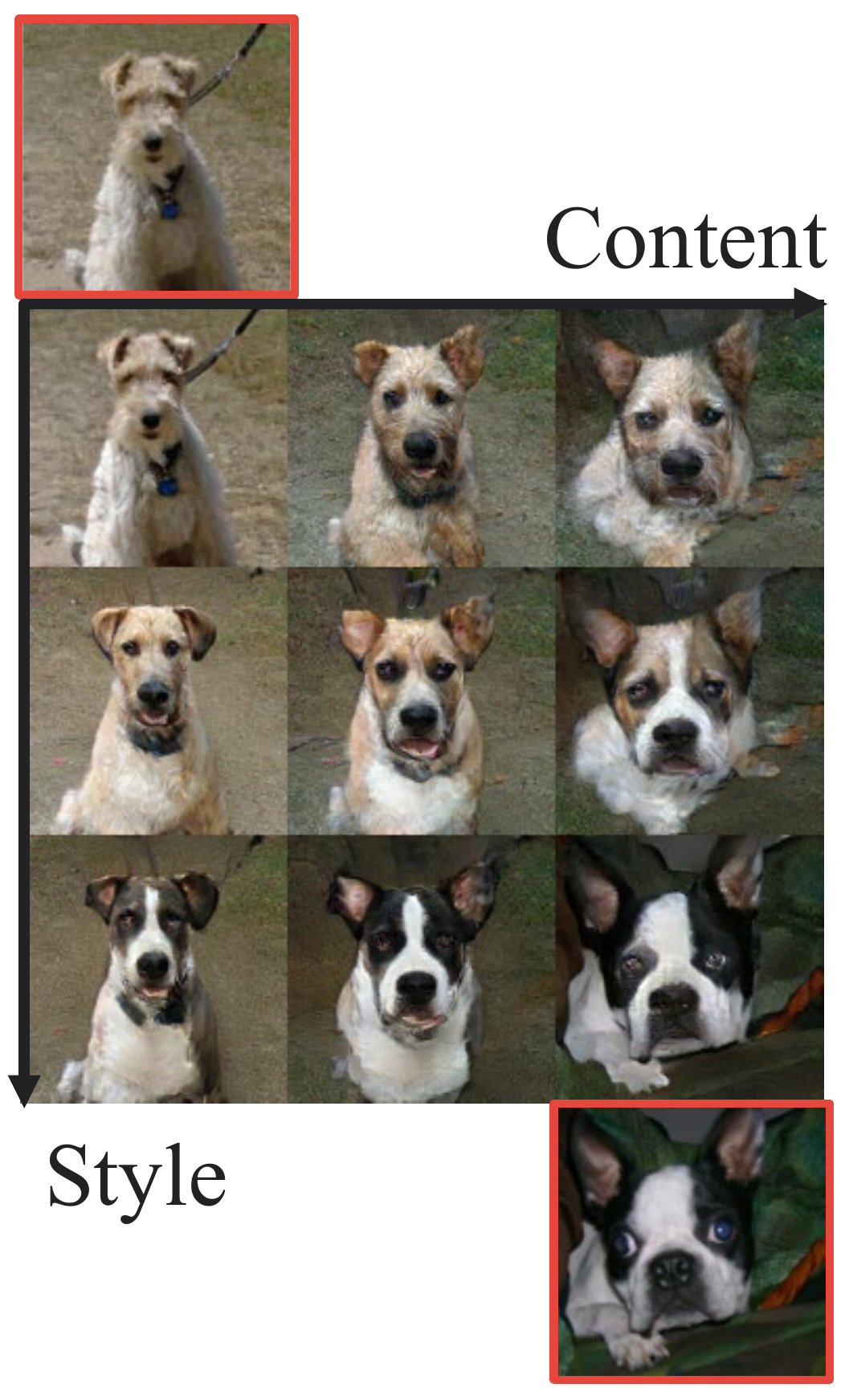}
    \caption{PatchGAN Discriminator}
    \label{fig:abl_patch_gan}
\end{subfigure}
\begin{subfigure}{0.24\linewidth}
    \includegraphics[width=\linewidth]{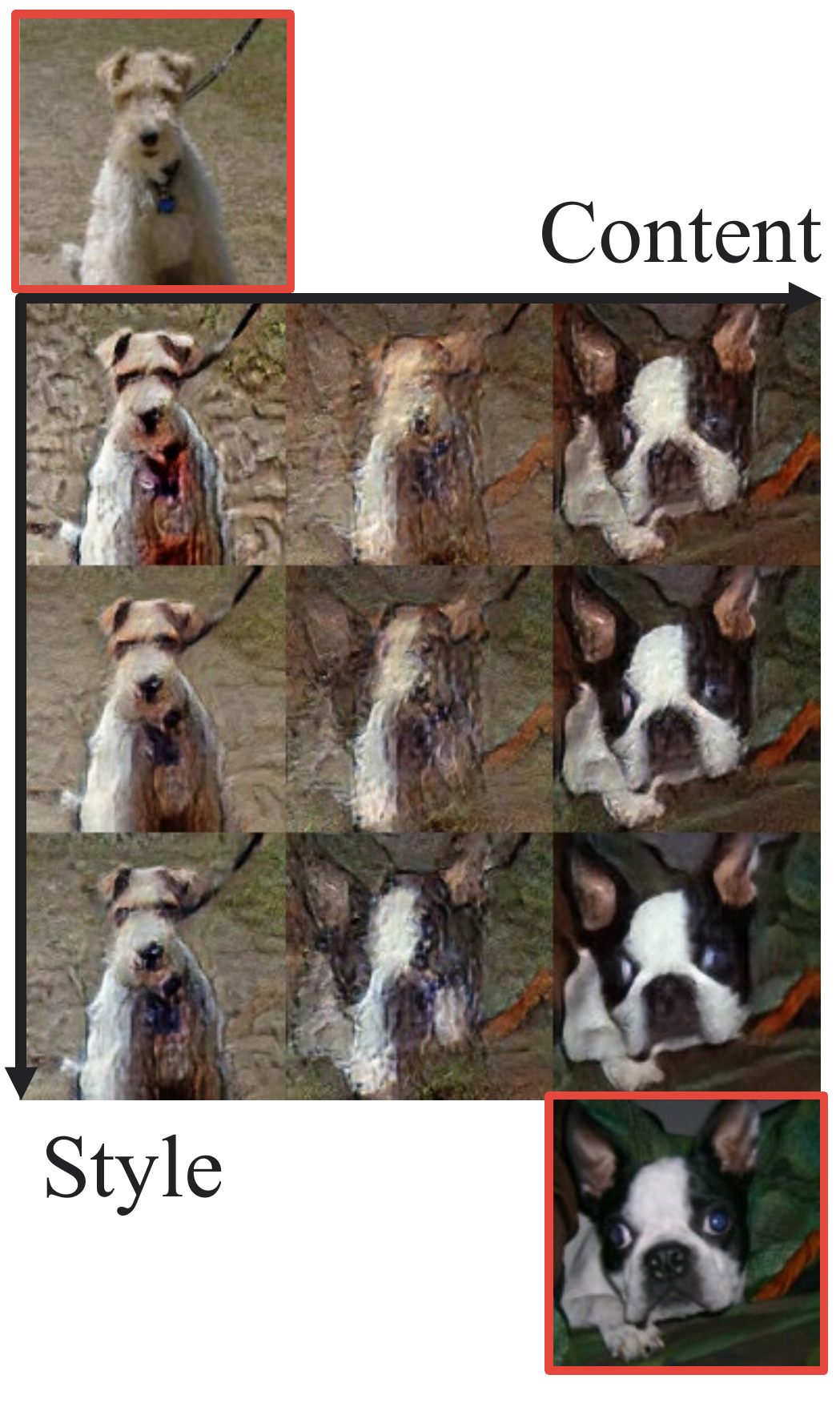}
    \caption{AC-GAN Discriminator}
    \label{fig:abl_ac_gan}
\end{subfigure}
\begin{subfigure}{0.24\linewidth}
    \includegraphics[width=\linewidth]{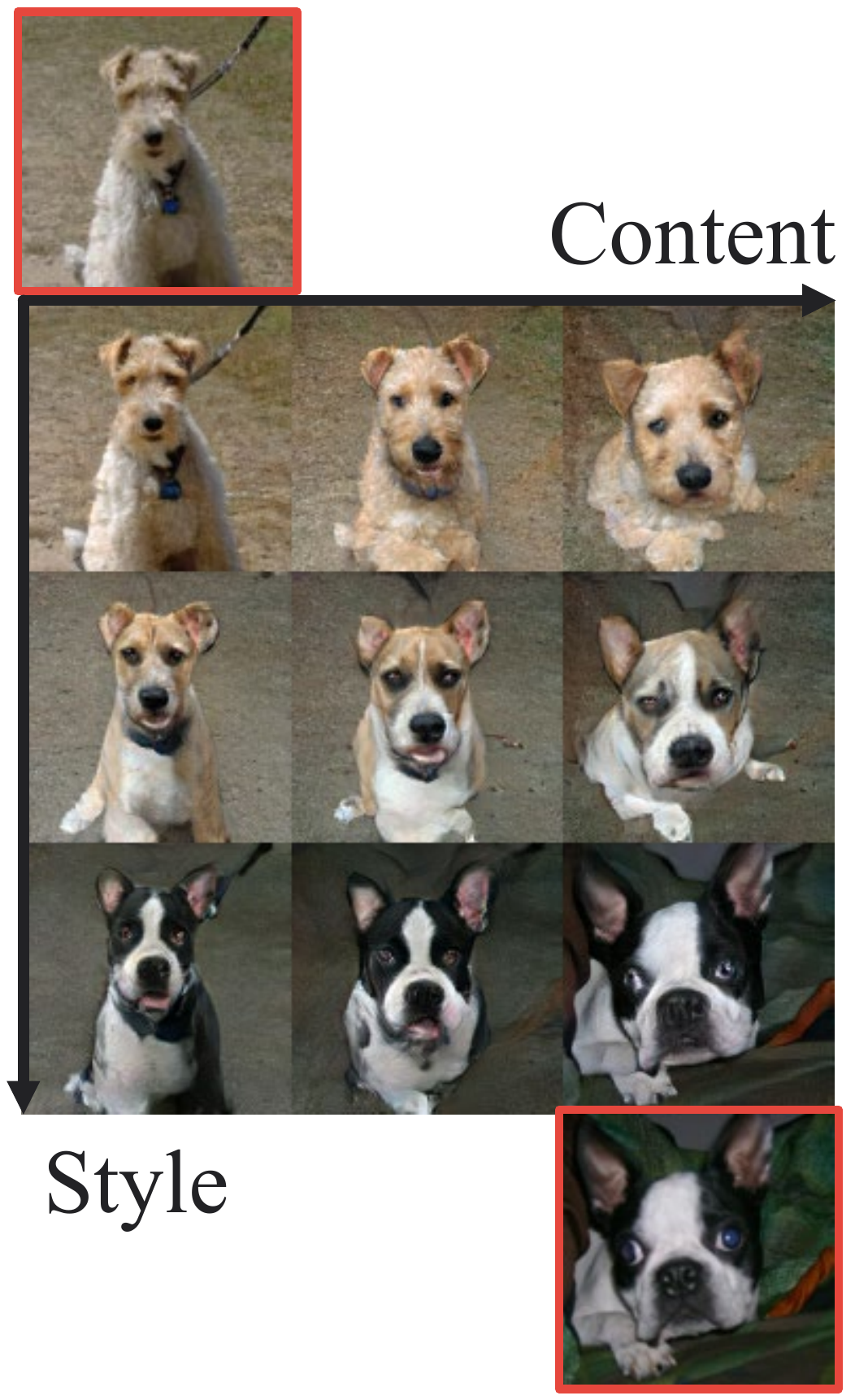}
    \caption{Ours}
    \label{fig:abl_full}
\end{subfigure}
\caption{An ablation test. Each 3x3 grid image represents a visualization of the 2D content-style manifold produced by each ablation setup with $\tau = 0.3$. The red boxes indicate the input images. }
\label{fig:ablation}
\end{figure*}

\subsection{Ablation Test}
\begin{table}[]
\centering
\caption{Quantitative results from a loss ablation test and an experiment on performance according to the discriminator type. The best result in each metric is in \textbf{bold}; the second best is \ul{underlined}. $\mathcal{L}^{idt}$, $\mathcal{L}^{mrp}$, $\mathcal{L}^{swp}$, and $\mathcal{L}^{cyc}$ represent the identity loss, the morphing loss, the swapping loss, and the cycle-swapping loss described in Section \ref{sec:objectives}, respectively. }
\label{tab:ablation}
\centering
\resizebox{\linewidth}{!}{%
\begin{tabular}{@{}l||r||rrrr@{}}
\hline 
    & \multicolumn{1}{c||}{Morphing}    & \multicolumn{4}{c}{Reconstruction}     \\ 
    & FID $\downarrow$   & MSE $\downarrow$ & PSNR $\uparrow$ & SSIM $\uparrow$ & LPIPS $\downarrow$ \\ \hline
w/o $\mathcal{L}^{idt}$ & 66.50      & 0.029      & 23.79      & 0.63      & 0.11 \\
w/o $\mathcal{L}^{mrp}$ & \ul{61.06} & 0.026      & \bf{26.53} & \bf{0.74} & \ul{0.08} \\
w/o $\mathcal{L}^{swp}$ & 63.35      & \bf{0.009} & \ul{26.25} & \ul{0.73} & \bf{0.07} \\
w/o $\mathcal{L}^{cyc}$ & 61.91      & 0.062      & 25.61      & 0.70      & 0.09 \\ \hline
PatchGAN Dis.           & 65.48      & \ul{0.014} & 25.36      & 0.68      & 0.09 \\
AC-GAN Dis.             & 181.20     & 0.017      & 21.95      & 0.50      & 0.20 \\ \hline
Ours                    & \bf{49.67} & 0.024      & 23.72      & 0.63      & 0.11 \\ \hline
\end{tabular}%
}
\end{table}

\textbf{Loss Ablation}
We performed an ablation test to analyze the impact of each objective loss. Table \ref{tab:ablation} shows that removing each of the four losses adversely affects the morphing performance of the network. A related qualitative assessment of the objectives is given in Figure \ref{fig:ablation}. While the elimination of the morphing loss resulted in a slight increase of the FID score from the full setup, Figure \ref{fig:abl_no_mrp} reveals that the morphing loss plays a crucial role and the network cannot learn a semantic change properly without it. For example, the top row clearly shows a wrong change in the style of the dog along the content axis. 

In the reconstruction evaluation, the absence of the morphing loss and the swapping loss leads to the best performance and the second-best performance. This is because morphing loss $\mathcal{L}^{mrp}$ and swapping loss $\mathcal{L}^{swp}$ are directly related to the morphing effect and there is a trade-off between learning a morphing transition and restoring the input image. 
Despite a somewhat mid-level quantitative performance for reconstruction, the  visual quality of the restored input images from the full setup is not much different from that of the images reconstructed with the elimination of each term.
On the other hand, the full setup results show more reasonably interpolated images compared with the results from other setups.
For example, the content transition presented in each row in Figure \ref{fig:abl_full} shows that the preservation of each dog's style is compared well with other results presented in the corresponding rows in Figures \ref{fig:abl_no_idt}-\ref{fig:abl_no_cyc}.

\textbf{Discriminator}
We replaced our discriminator $D$ with the PatchGAN discriminator \cite{isola2017image} and the AC-GAN discriminator \cite{odena2017conditional} in our network while maintaining other conditions to analyze the effect of each type of discriminator. Similar to the loss elimination experiment, the trade-off between a morphing ability and an input image reconstruction is evident as shown in Table \ref{tab:ablation}. In this case again, the qualitative evaluation in Figures \ref{fig:abl_patch_gan}-\ref{fig:abl_full} verify that the lack in our reconstruction performance leads to the insignificant visual difference in the results, while other discriminators produce significantly inferior morphing results compared with ours. The use of the AC-GAN discriminator prevents the network from learning the basic morphing transition at all, and the PatchGAN also adversely affects the quality of the disentangled transition  compared with the full setup that uses a multi-task discriminator.


\subsection{Comparison}

\begin{figure*}[!h]
\includegraphics[width=\linewidth]{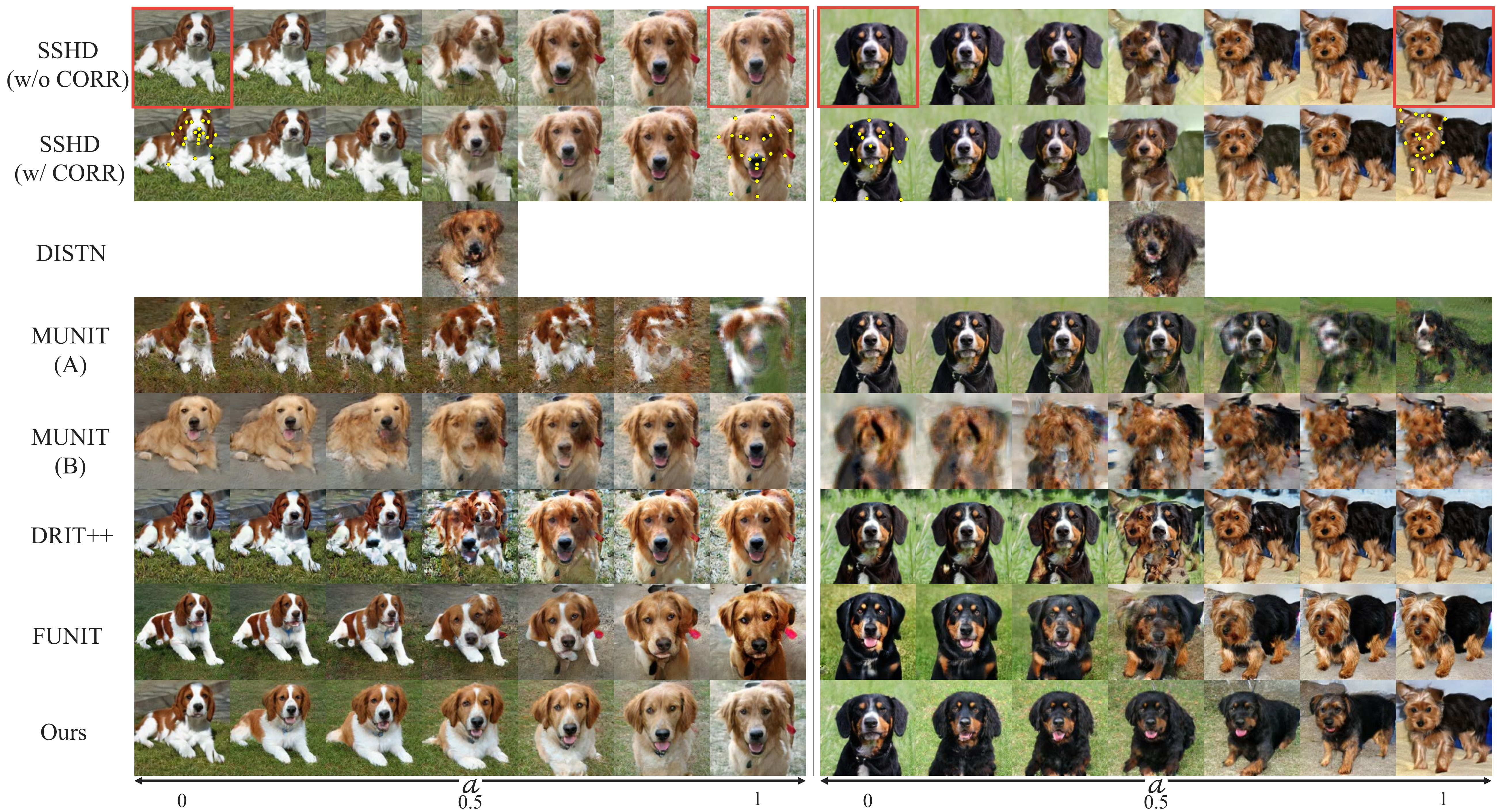}
\caption{Qualitative comparison. \rvs The red boxes in the first row indicate the input images for all of the methods. \stoprvs The yellow dots represent user correspondence points used in the SSHD method. MUNIT (A) and (B) are the results from the decoder responsible for the style of the first and the second input image, respectively. Therefore, the results from MUNIT show only content transition in each row. }
\label{fig:comparison}
\end{figure*}

\textbf{Baselines}
We compared the results from our method with those from three leading baseline methods. The first one is the conventional image morphing method that utilizes structural similarity in a halfway domain (SSHD) \cite{liao2014automating}. The second one is the approach that distills the pre-trained network, which is similar to the method used by our approach. \citet{viazovetskyi2020stylegan2} trained the Pix2pixHD model \cite{wang2018high} by feeding a pair of images to output a single intermediate image that has a morphing effect (DISTN). In the same way, we trained the Pix2pixHD model with our BigGAN data triplets\footnotemark $\{x_{A},x_{B},x_{0.5}\}$ for fair comparison. The third one is the image-to-image translation method such as MUNIT \cite{huang2018munit}, DRIT++ \cite{lee2019dritpp}, and FUNIT \cite{liu2019funit}. Similar to our network, these image translation models map images to style and content codes. DRIT++ and FUNIT were trained as the authors had proposed, and DRIT++ accepted an additional input class code along with the input images from the dataset. Since MUNIT translates between two classes only, we assign two classes to train MUNIT. \rvs For a fair visual comparison, we retrained the model from scratch with our BigGAN dataset. At the inference stage, we linearly interpolated the style and content code of the given pair of images by adjusting $\alpha$ as done in our method. \stoprvs
For all of the baselines codes, we utilized the implementation provided by the authors.

\footnotetext{In Equation \ref{eq:data}, the value of $\alpha$ is fixed at 0.5 to build a set of data triplets.}

\begin{table}[]
\caption{Quantitative comparison. The best result in each metric is in \textbf{bold}; the second best is \ul{underlined}.}
\label{tab:comparison}
\centering
\resizebox{\linewidth}{!}{%
\begin{tabular}{@{}l||r||rrrr@{}}
\hline
    & \multicolumn{1}{c||}{Morphing}    & \multicolumn{4}{c}{Reconstruction}     \\ 
    & FID $\downarrow$ & MSE $\downarrow$ & PSNR $\uparrow$ & SSIM $\uparrow$ & LPIPS $\downarrow$ \\ \hline
DISTN  & 126.85     & -         & -          & -         & -           \\
DRIT++ & 54.59      & \ul{0.06} & \ul{22.45} & \bf{0.77} & \bf{0.08} \\
FUNIT  & \ul{50.47} & 0.15      & 12.75      & 0.16      & 0.32        \\
Ours   & \bf{49.67} & \bf{0.02} & \bf{23.72} & \ul{0.63} & \ul{0.11} \\ \hline
\end{tabular}%
}
\end{table}

\textbf{Quantitative Comparison}
We excluded the results from MUNIT because it can only learn two classes and cannot handle many breeds of dogs included in AFHQ.
Table \ref{tab:comparison} shows that our method outperforms all of the baseline methods. Our method achieved the best performance in FID, MSE, and PSNR, and the second best performance in SSIM and LPIPS.

\textbf{Qualitative Results}
We qualitatively compared the results from our method with those from the baseline methods. To visualize the morphing ability of the baseline methods, we varied the value of $\alpha$ from $0$ to $1$ as performed in our method. Figure \ref{fig:comparison} shows the morphing results. It is apparent that the combination of warping and blending cannot handle the large occlusion implied in two input images, creating strong ghosting artifacts in the mid point of the morphing regardless of use of the correspondences as evidenced by the results from SSHD.
The mid point output produced by DISTN appears to be a somewhat reasonable mixture of the two input images. Unfortunately, however, the quality of the output image is low.
Although the methods for image translation, such as DRIT++ and FUNIT, have  been successful in style interpolation that mainly changes the appearance of an object, they are not capable of generating a semantic morphing transition that requires both content and style changes. The right hand side result from FUNIT shows a transition from one class to another, yet the mid point image still appears to be a physical blending of the two images instead of a semantic blending. In addition, FUNIT tends to alter the original input images a lot more than the other methods do. DRIT++ preserves the original input images well, but the mid point image is more like a superimposed version of the two input images.
This visual interpretation is consistent with the quantitative evaluation reported in Table \ref{tab:comparison}.

Because MUNIT assumes that the style space is decoupled for every class, it is impossible to change the style between the two input images. In addition, like other translation models it was not able to morph the geometry of the content successfully.
In contrast to all of these baseline methods, our method can handle both content and style transfer given significantly different geometrical objects that contain 
occluded regions and produce visually pleasing morphing effects. \rvs In particular, as the value of $\alpha$ changes, the pose and appearance of the dog change semantically smoothly. This shows a sharp contrast with the most of the baseline methods, which produce a superimposed image only at $\alpha = 0.5$ with no continuous semantic changes.

\begin{figure}[!h]
    \includegraphics[width=\linewidth]{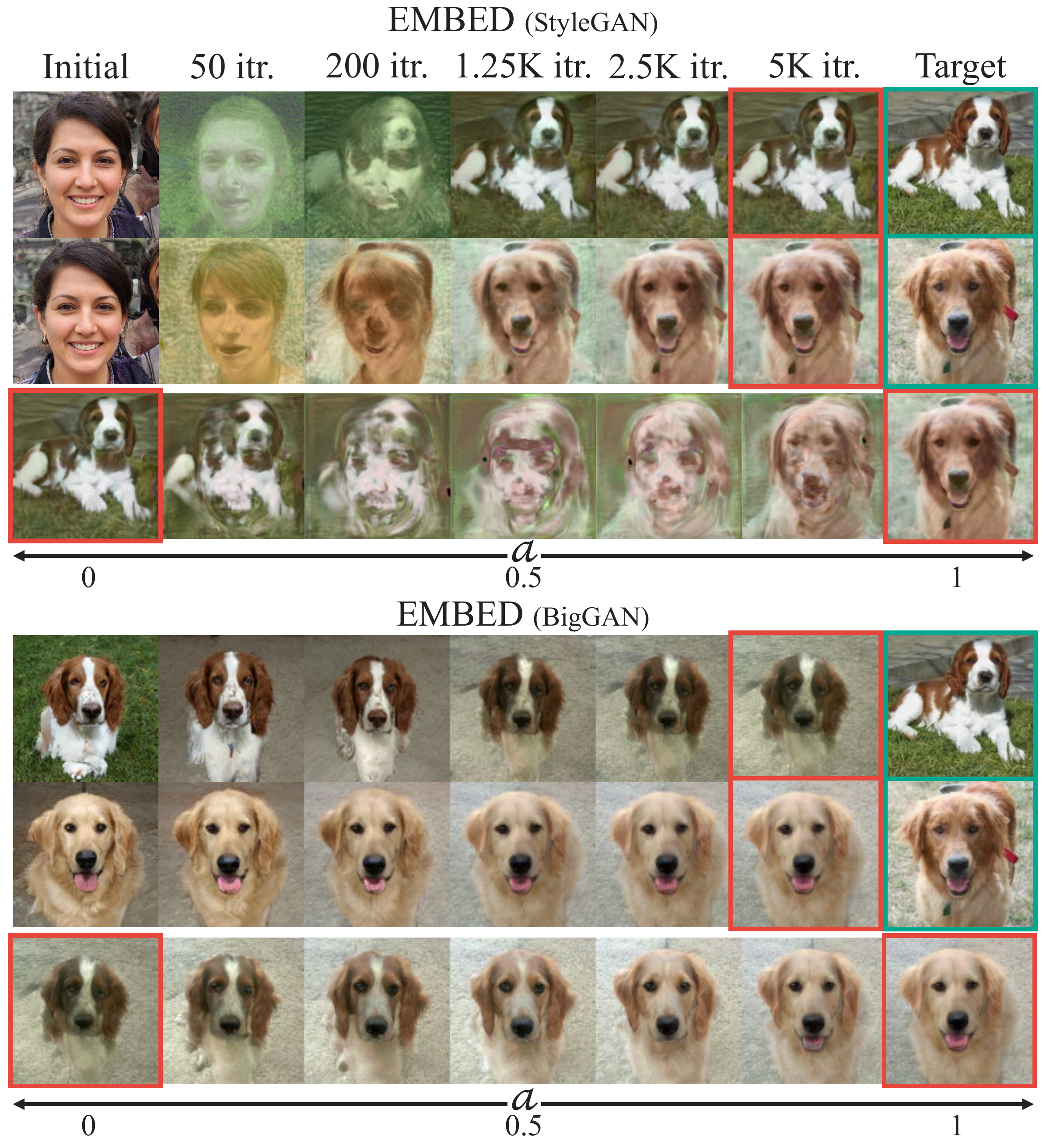}
    \caption{\rvs Results from embedding algorithms. The green boxes indicate the target images to be embedded. The red boxes indicate the final embedding results. \stoprvs}
    \label{fig:comparison_embed}
\end{figure}

\textbf{Comparison with Embedding Methods}
We additionally performed experiments to qualitatively compare the results from our method and the state-of-the-art embedding algorithm \cite{abdal2019image2stylegan}. As \citet{abdal2019image2stylegan} suggested, we used a gradient descent algorithm \cite{creswell2018inverting} to embed a given image onto the manifold of the pre-trained StyleGAN \cite{karras2019style} with the FFHQ dataset (EMBED \scriptsize (StyleGAN)\normalsize). The optimization steps and hyperparameters of the optimizer were set as in EMBED \scriptsize (StyleGAN)\normalsize. The first and the second rows in Figure \ref{fig:comparison_embed} show the process of embedding the input images of the first example in Figure \ref{fig:comparison} into the latent space of the StyleGAN. After 5000 iterations, the embedding result reasonably converged to the target image as shown in the images in the red box. However, when we performed a morphing process by interpolating the two latent vectors associated with the two images, the result was not at all satisfactory, as demonstrated in the third row. This might have been because the StyleGAN is currently trained only with face data.

For a fairer comparison, we used the pre-trained BigGAN (EMBED \scriptsize (BigGAN)\normalsize) for embedding. In the initial experiment, the embedding algorithm failed to simultaneously optimize latent vector $z$ and conditional class embedding vector $e$ of the BigGAN. Therefore, we initialized $e$ manually according to the target image and optimized only $z$. The results are shown in the fourth and fifth rows in Figure \ref{fig:comparison_embed}. As the sixth column reveals, the embedding results were not satisfactory even after 5000 iterations. For example, the pose of the spaniel is completely lost in the fourth image while the appearance details of the retriever are not fully recovered in the fifth image. Given the embedding results, EMBED \scriptsize (BigGAN)\normalsize \text{ }faithfully produced a morphing effect as shown in the sixth row.

\begin{figure*}[!h]
    \includegraphics[width=\linewidth]{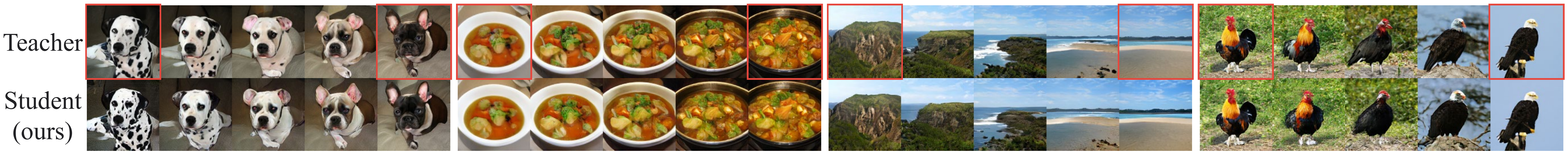}
    \caption{\rvs Comparison of the results from the teacher network and from the student network. The images in the red boxes created by the teacher network are used as the input image for our generator. \stoprvs}
    \label{fig:comparison_teacher}
\end{figure*}

\textbf{Comparison with the Teacher Network}
To verify how well our student network can mimic the ability of semantic transformation of the teacher network, we compared the results from our generator with the interpolation results in the latent space of the teacher generative model. The comparisons performed on \textsc{Dogs}, \textsc{Foods}, \textsc{Landscapes}, and \textsc{Birds} are shown in Figure \ref{fig:comparison_teacher}. The morphing sequence was obtained by inputting the two end images created from the teacher network into the student generator. It is clear that the quality of the morphed image produced by the generator was comparable to that produced by the interpolation in the teacher network.

We report the time spent to generate a morphing image by various methods in Table \ref{tab:timing}. Our method was much more efficient than the conventional methods that require optimization for the warping functions and the embedding methods that require iterative gradient descent steps.

\begin{table}[]
\caption{\rvs Timing statistics \stoprvs}
\label{tab:timing}
\centering
\resizebox{\linewidth}{!}{%
\begin{tabular}{@{}lcccccc@{}}

\hline
                                            & Device & \multicolumn{5}{c}{Computation time (s)} \\ \hline
\multicolumn{2}{c|}{Optimization steps (itr.)}   & 50 & 200 & 1.25K & 2.5K & 5K \\ \hline
\multicolumn{1}{l||}{EMBED (StyleGAN)}      & \multicolumn{1}{c|}{GPU} & 16.72 & 60.80 & 372.96 & 747.31 & 1,495.66 \\
\multicolumn{1}{l||}{EMBED (BigGAN)}        & \multicolumn{1}{c|}{GPU} & 11.00 & 39.98 & 244.61 & 488.13 & 973.46 \\ \hline
\multicolumn{1}{l||}{SSHD w/o CORR}         & \multicolumn{1}{c|}{CPU} & \multicolumn{5}{c}{4.21}       \\
\multicolumn{1}{l||}{SSHD w/ CORR}          & \multicolumn{1}{c|}{CPU} & \multicolumn{5}{c}{3.43}       \\ \hline
\multicolumn{1}{c||}{Ours}                  & \multicolumn{1}{c|}{CPU} & \multicolumn{5}{c}{\bf{1.26}}     \\
\multicolumn{1}{c||}{Ours}                  & \multicolumn{1}{c|}{GPU} & \multicolumn{5}{c}{\bf{0.12}}     \\ \hline

\end{tabular}%
}
\end{table}

\stoprvs

\section{Extensions and Applications}

\subsection{Multi Image Morphing}
\begin{figure*}[h]
\begin{subfigure}{0.32\linewidth}
    \includegraphics[width=\linewidth]{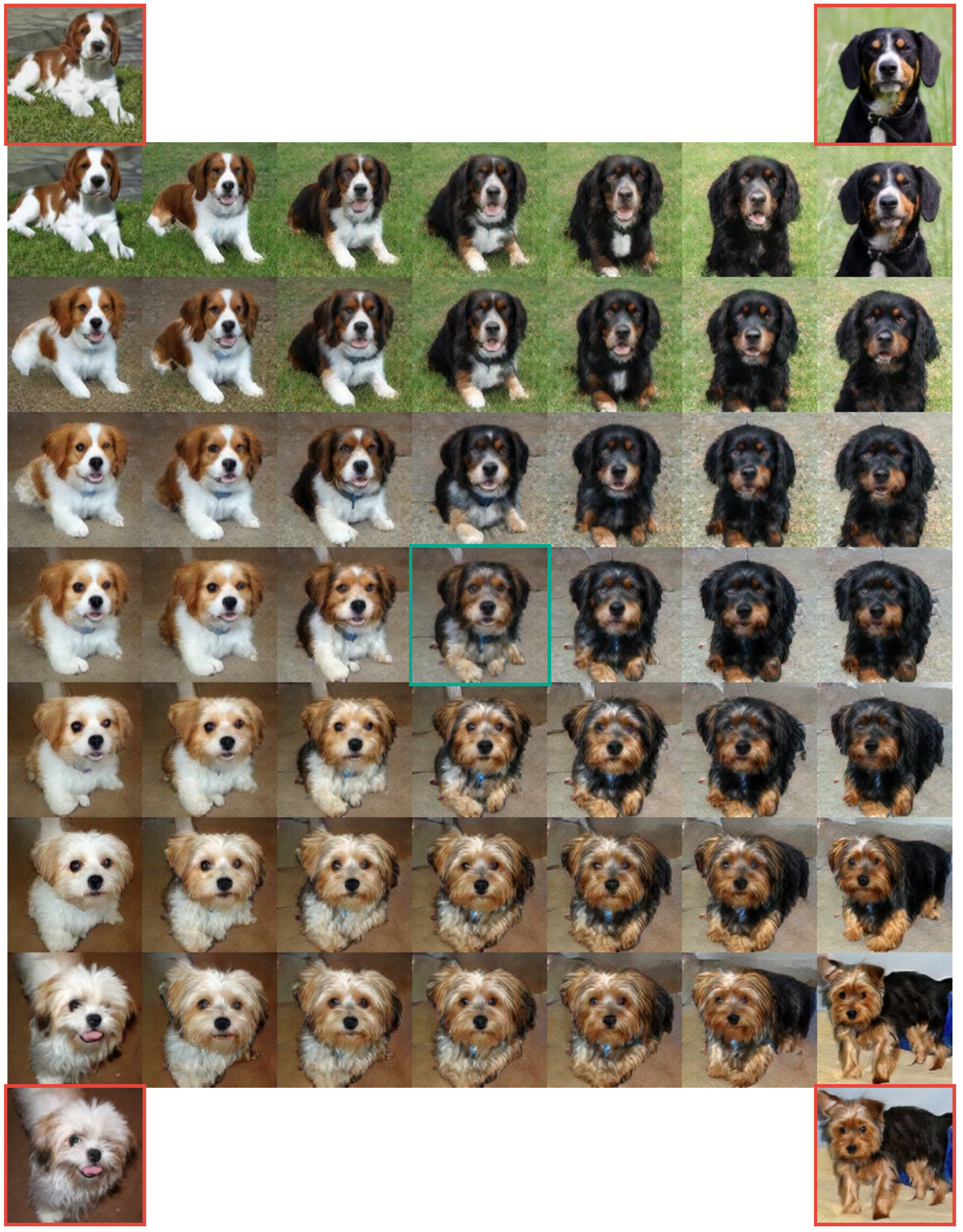} 
    \caption{Basic transition}
    \label{fig:mm1}
\end{subfigure}
\begin{subfigure}{0.32\linewidth}
    \includegraphics[width=\linewidth]{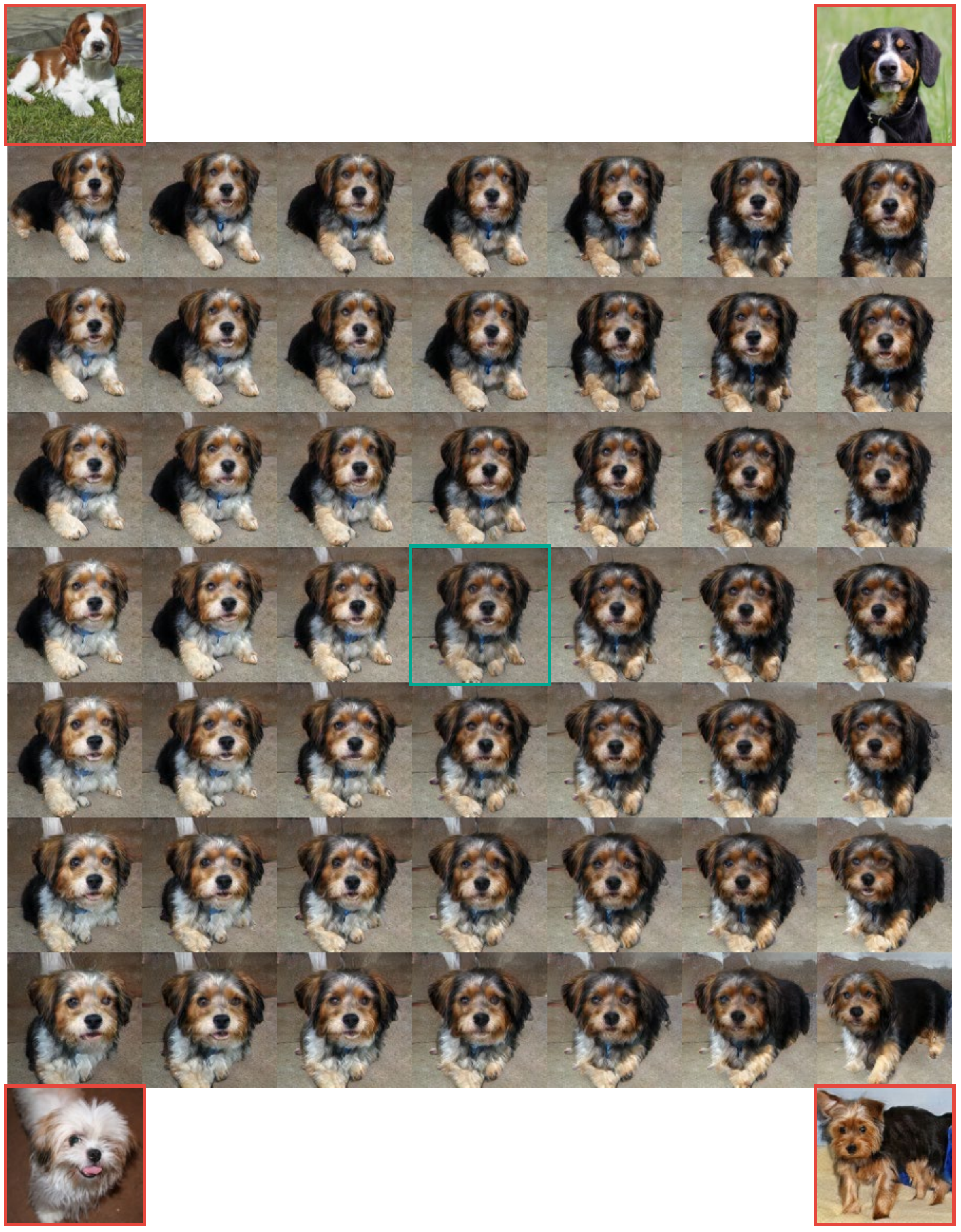}
    \caption{Disentangled transition (content)}
    \label{fig:mm2}
\end{subfigure}
\begin{subfigure}{0.32\linewidth}
    \includegraphics[width=\linewidth]{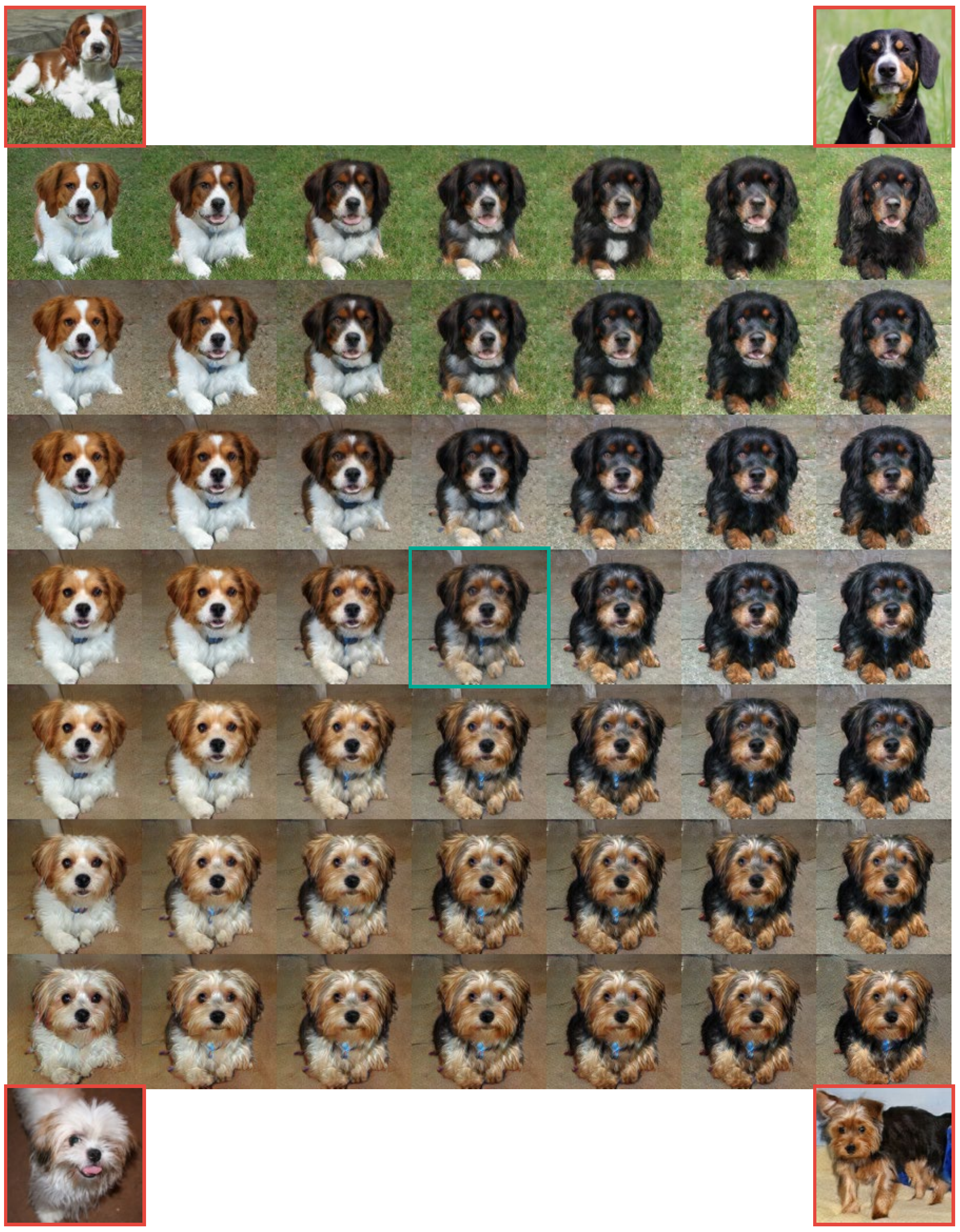}
    \caption{Disentangled transition (style)}
    \label{fig:mm3}
\end{subfigure}
\caption{Multi image morphing. The red boxes indicate the reference images. The green box shows the intermediate image produced when all of the weights are equal. (a) The basic transitions between four images. (b) The disentangled transitions where the content is interpolated between four images while all of the style weights $w_{s_{1,2, \ldots, M}}$ are fixed at $\frac{1}{M}$. (c) The disentangled transitions where the style is interpolated between four images while all of the content weights $w_{c_{1,2, \ldots, M}}$ are fixed at $\frac{1}{M}$. }
\label{fig:multi_morphing}
\end{figure*}

So far, we have explained morphing given two input images, but Neural Crossbreed framework is not limited to handling only two images. The generalization to handle multiple images becomes possible by replacing interpolation parameters $\alpha_c$ and $\alpha_s$ used in Equation \ref{eq:factor1} with the weights for $M$ images. The modification can be expressed as follows:
\begin{align} \label{eq:multi}
    \begin{split}
    y_{w_{c}w_{s}} &= G(x_{1,2, \ldots, M},w_{c_{1,2, \ldots, M}}, w_{s_{1,2, \ldots, M}}), \\
    &= F(c_{w_c}, s_{w_s}) , \\
    \text{where } c_{w_c} &= \sum_{m=1}^{M} w_{c_m} E_c(x_m)\text{, } s_{w_s} = \sum_{m=1}^{M} w_{s_m} E_s(x_m).
    \end{split}
\end{align}
Here, content weight $w_{c_{1,2, \ldots, M}}$ and style weight $w_{s_{1,2, \ldots, M}}$ satisfy the following constraints: $\sum_{m=1}^{M}w_{c_m} = 1 \text{ and } \sum_{m=1}^{M}w_{s_m} = 1$. Note that this adjustment is applied only at runtime using the same network, without any modification to the training procedure. Figure \ref{fig:multi_morphing} shows an example of the basic transition and the disentangled transition between four input images.

\subsection{Video Frame Interpolation}
\begin{figure*}[h!]
\includegraphics[width=\linewidth]{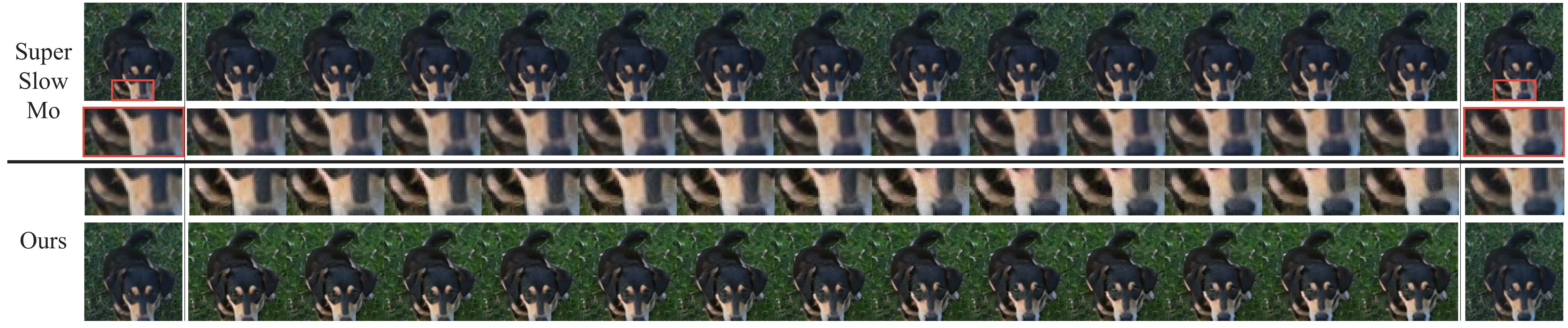}
\caption{ \rvs Frame interpolation by our method and comparison with the result from SuperSlowMo. \stoprvs The leftmost and the rightmost images are two adjacent frames in the video. The intermediate images are generated by the neural network with the two adjacent frames as input.}
\label{fig:frame_interp}
\end{figure*}

Assuming that the frames are closely related and share a similar camera baseline, frame interpolation produces in-between images given two neighbor frames. Because Neural Crossbreed aims to handle significantly different images, we can consider frame interpolation as a sub-problem of image morphing and produce a high frame rate video. Figure \ref{fig:frame_interp} shows intermediate frames created by inputting two neighbor frames into the generator. \rvs Our method does not enforce temporal coherency explicitly and therefore such a property is not guaranteed for all categories. Nonetheless, for puppy images, the results were qualitatively similar to those from a recent frame interpolation study (SuperSlowMo) \cite{jiang2018super} in a preliminary experiment. \stoprvs Specifically, a smooth change of the fur pattern and the position of the snout is apparent in the enlarged area. Note that we can create as many interpolated frames as we want by changing the value of alpha. A comparison with the results from the conversion from the standard to a high frame rate video can be seen in the accompanying video.

\subsection{Appearance Transfer}

\begin{figure}[h!]
    \includegraphics[width=\linewidth]{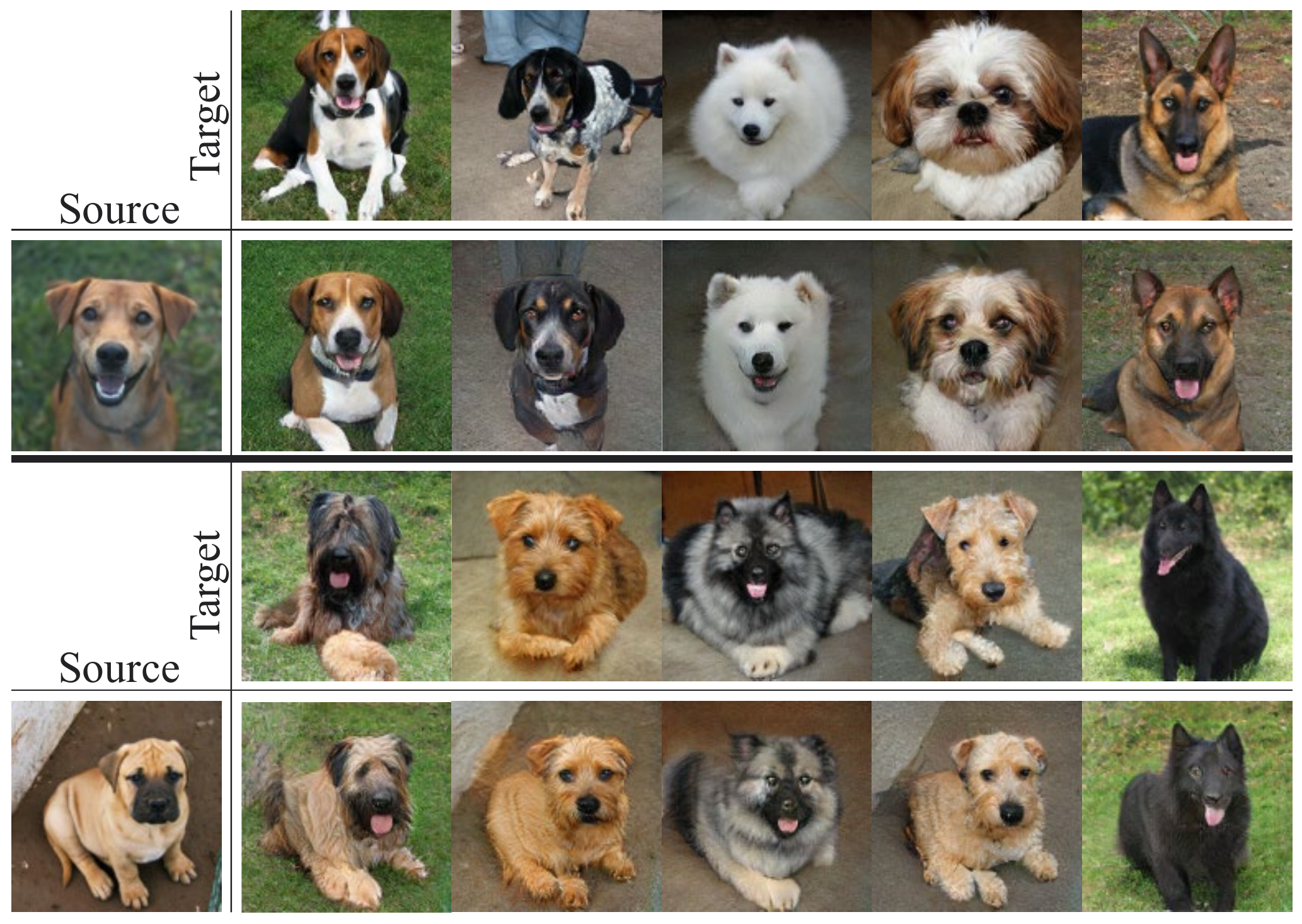}
    \caption{Appearance transfer. The first column is the source image to be manipulated. The first and third rows contain target images with a style that will be applied to the source image. The rest are the images from the generator with the source and the target images as input.}
    \label{fig:style_transfer}
\end{figure}

Disentangled outputs from the generator include images with the style and content components of the input image swapped. Therefore, just like any other image translation method, Neural Crossbreed can also be applied to create images with altered object appearances. Figure \ref{fig:style_transfer} shows the results from converting the source image to have the style of a target image.


\section{Discussion and Future Work}

\begin{figure}[h!]
  \includegraphics[width=\linewidth]{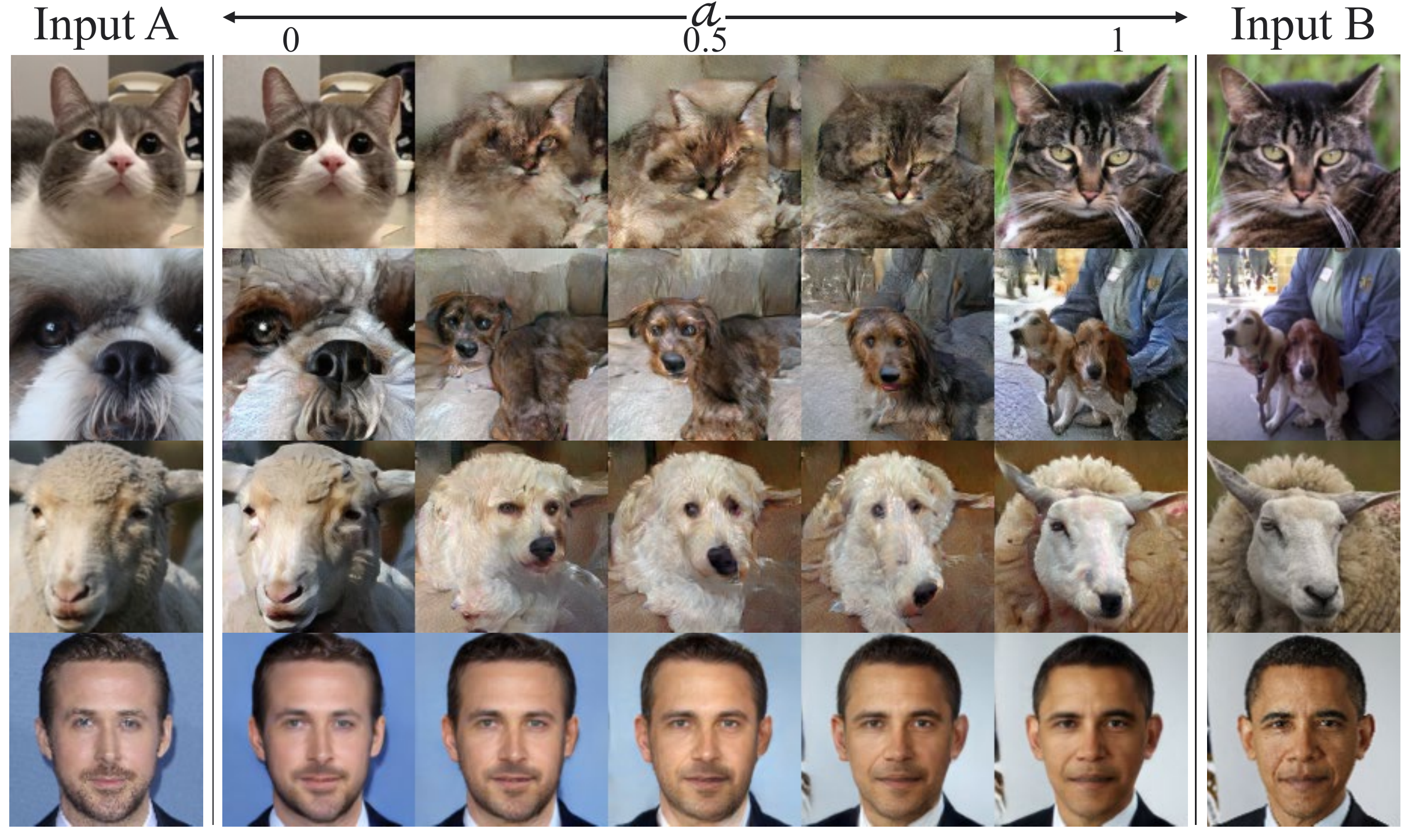}
  \caption{\rvs Cases of failure. Note that the first row used the feline data from the BigGAN to train the generator, the second and third rows used the \textsc{Dogs} data from the BigGAN, and the bottom row used the human face data from the StyleGAN2. The top row shows a case of failure caused by images with relatively low quality obtained from the BigGAN. The second row shows a case of failure caused by an extreme pose and the presence of multiple objects. The third row shows a case of failure caused by a category not contained in the training dataset. The bottom row shows a case of failure caused by a single class generative model as a teacher network.
  \stoprvs}
  \label{fig:fail}
\end{figure}

Although Neural Crossbreed can produce a high quality morphing transition between input images, it also has some limitations. \stoprvs Figure \ref{fig:fail} shows examples of the failure cases.
The image classes that can be morphed are limited to the categories generated by a teacher network, and if the teacher network cannot generate high-quality images in the category, an example being feline animals, the network fails to learn semantic changes in that domain.
\rvs The generator also failed to handle images that deviate too much from the training data distribution, such as the images that contain extreme object poses, multiple objects, or objects of a category not learned in the training phase.

We performed a preliminary test to morph human faces \rvs and found that the resulting images were blurry. \stoprvs For the generation of human faces for training, we tried StyleGAN2 \cite{karras2019analyzing}, as a teacher network. Because StyleGAN2 is not a class conditional GAN and manipulating human facial attributes are more suitable for a multi-label problem than a multi-class problem, we were not able to successfully apply StyleGAN2 directly to our framework where a multi-task discriminator plays an important role for the morphing performance (See Table \ref{tab:ablation}). It will be an interesting direction to investigate a new discriminator for the multi-label problem that works in the context of image morphing. We will look into this issue in the future.

\rvs
In this study, we focused only on morphing of a foreground object and did not pay attention to the background. In the future, we would also like to consider the background by integrating a pre-trained semantic segmentation network \cite{long2015fully} or a self-attention component \cite{zhang2019self} into our framework. While Figure \ref{fig:comparison} shows that the morphing effect of our method is superior to that by the baseline methods, we acknowledge that there is room for improvement in the absolute image quality. For this, through an additional refinement network or a custom post-process can be utilized. We will leave this as another important future work.
\stoprvs

\section{Conclusion}
Unlike conventional approaches, our proposed Neural Crossbreed can morph input images with no correspondences explicitly specified; it also can handle objects with significantly different poses by generating semantically meaningful intermediate content using a pair of teacher-student generators. In addition, content and style transitions are effectively disentangled in a latent space to provide the user with a tight control over the morphing results.

We proposed the first feed-forward network that learns semantic changes from the pre-trained generative model for an image morphing task. Our network exploits knowledge distillation where a student network learns from a teacher network to perform the same task with comparable or better performance.
Our student network, whose purpose is to morph an image into another, learns a solution from the teacher network that maps a latent vector to an image. This concept is similar to \textit{\textbf{``analogous inspiration''}} that helps a network widen its knowledge and discover new perspectives (image morphing) from other contexts (mapping between a latent space and realistic images).
We hope that our work will inspire similar approaches for conventional image or video domain problems in the era of deep learning.








\bibliographystyle{ACM-Reference-Format}
\bibliography{99_bib}

\appendix

\begin{table}[]
\caption{\rvs Overview of the network architecture. Convolutional filters are specified in the format of "k(\#kernel size)s(\#stride)”. $H$ and $W$ indicate the height and the width of input image $x$. \stoprvs}
\label{tab:gen_arc}
\centering
\resizebox{\linewidth}{!}{%
\begin{tabular}{@{}llcccc@{}}
\hline
Content Encoder $E_c$  & Filter & Act. & Norm. & Down  & Output               \\ \hline
Conv    & k7s1   & ReLU & -     & -       & $32 \times H\times W$       \\
PResBlk & k3s1   & ReLU & IN    & AvgPool & $64 \times H/2 \times W/2$ \\
PResBlk & k3s1   & ReLU & IN    & AvgPool & $128 \times H/4 \times W/4$ \\
PResBlk & k3s1   & ReLU & IN    & AvgPool & $256 \times H/8 \times W/8$ \\

PResBlk & k3s1   & ReLU & IN    & -       & $256 \times H/8 \times W/8$ \\
PResBlk & k3s1   & ReLU & IN    & -       & $256 \times H/8 \times W/8$ \\
PResBlk & k3s1   & ReLU & IN    & -       & $256 \times H/8 \times W/8$ \\ \hline

               &        &      &       &       &                         \\ \hline
Style Encoder $E_s$ & Filter & Act. & Norm. & Down  & Output                 \\ \hline
Conv    & k7s1   & ReLU & IN    & -       & $32 \times H\times W$       \\
PResBlk & k3s1   & ReLU & IN    & AvgPool & $64 \times H/2 \times W/2$ \\
PResBlk & k3s1   & ReLU & IN    & AvgPool & $128 \times H/4 \times W/4$ \\
PResBlk & k3s1   & ReLU & IN    & AvgPool & $256 \times H/8 \times W/8$ \\
PResBlk & k3s1   & ReLU & IN    & AvgPool & $512 \times H/16 \times W/16$ \\
PResBlk & k3s1   & ReLU & IN    & AvgPool & $1024 \times H/32 \times W/32$ \\
PResBlk & k3s1   & ReLU & IN    & AvgPool & $2048 \times H/64 \times W/64$ \\
\multicolumn{3}{l}{Global Sum Pooling} & &  & $2048 $                         \\ \hline

               &        &      &       &       &                         \\ \hline
\multicolumn{2}{l}{Mapping Network}    & Act. &    &     & Output \\ \hline
\multicolumn{2}{l}{Linear}             & ReLU &    &     & $512$  \\
\multicolumn{2}{l}{Linear}             & ReLU &    &     & $512$  \\
\multicolumn{2}{l}{Linear}             & ReLU &    &     & $512$  \\
\multicolumn{2}{l}{Linear}             & ReLU &    &     & $512$  \\
\multicolumn{2}{l}{Linear}             & ReLU &    &     & $3968$ \\ \hline

        &        &      &       &       &                         \\ \hline
Decoder $F$ & Filter & Act. & Norm. & Up  & Output                 \\ \hline
PResBlk & k3s1   & ReLU & AdaIN & -       & $256 \times H/8 \times W/8$ \\
PResBlk & k3s1   & ReLU & AdaIN & -       & $256 \times H/8 \times W/8$ \\
PResBlk & k3s1   & ReLU & AdaIN & -       & $256 \times H/8 \times W/8$ \\

PResBlk & k3s1   & ReLU & AdaIN & Upsample & $128 \times H/4 \times W/4$ \\
PResBlk & k3s1   & ReLU & AdaIN & Upsample & $64 \times H/2 \times W/2$ \\
PResBlk & k3s1   & ReLU & AdaIN & Upsample & $32 \times H \times W$ \\
Conv    & k7s1   & Tanh & -     & -       & $3 \times H \times W$    \\ \hline

&        &      &       &       &                         \\ \hline
Discriminator $D$  & Filter & Act. & Norm. & Down  & Output               \\ \hline
Conv    & k7s1   & -     & -    & -       & $32 \times H \times W$       \\
PResBlk & k3s1   & LReLU & -    & -       & $32 \times H \times W$       \\
PResBlk & k3s1   & LReLU & -    & AvgPool & $64 \times H/2 \times W/2$   \\
PResBlk & k3s1   & LReLU & -    & -       & $64 \times H/2 \times W/2$   \\
PResBlk & k3s1   & LReLU & -    & -       & $128 \times H/2 \times W/2$  \\
Conv    & k7s1   & LReLU & -    & -       & $L \times H/2 \times W/2$    \\ \hline

\end{tabular}%
}
\end{table}

\rvs
\section{Architectural Details} \label{appx:architecture}

In Table \ref{tab:gen_arc}, we describe the layers of generator $G$ and discriminator $D$ in order. Except for the mapping network, which consists of five fully connected layers, most of the other networks employed pre-activated residual blocks (PResBlk) \cite{he2016identity}. Content encoder $E_c$ consists of three down-sampling layers and three bottleneck layers, while style encoder $E_s$ consists of six down-sampling layers and the final global sum-pooling layer. Therefore, content code $c$ and style code $s$ are interpolated in the form of a latent tensor and a latent vector, respectively. Decoder $F$ has a structure symmetrical to $E_c$ and contains three bottleneck layers and three up-sampling layers. In the layers of $F$, we use AdaIN to insert the style code into the decoder, whereas in the layers of both of the two encoders we use instance normalization (IN) \cite{ulyanov2017improved}. All of the activation functions in $G$ are the rectified linear unit (ReLU) \cite{nair2010rectified} except for the final hyperbolic tangent (Tanh) layer of decoder $F$, which produces an RGB image. Our discriminator $D$ predicts whether or not the image is real in $\abs{L}$ classes \cite{mescheder2018training,liu2019funit}, and each output uses a PatchGAN \cite{isola2017image} with dimensions of $H/2 \times W/2$. We use leaky ReLUs (LReLU) \cite{maas2013rectifier} with a slope of $0.2$ in $D$. 
Note that for content encoder $E_c$, style encoder $E_s$, and discriminator $D$, an RGB image (of size $3 \times H \times W$) is given as input, and content code $c$ (size $256 \times H/8 \times W/8$) and style code $s$ (size $2048$) are input to decoder $F$ and the mapping network, respectively.

\stoprvs

\section{Truncation Trick} \label{appx:truncation}

\begin{wrapfigure}{L}{0.4\linewidth}
    \includegraphics[width=\linewidth]{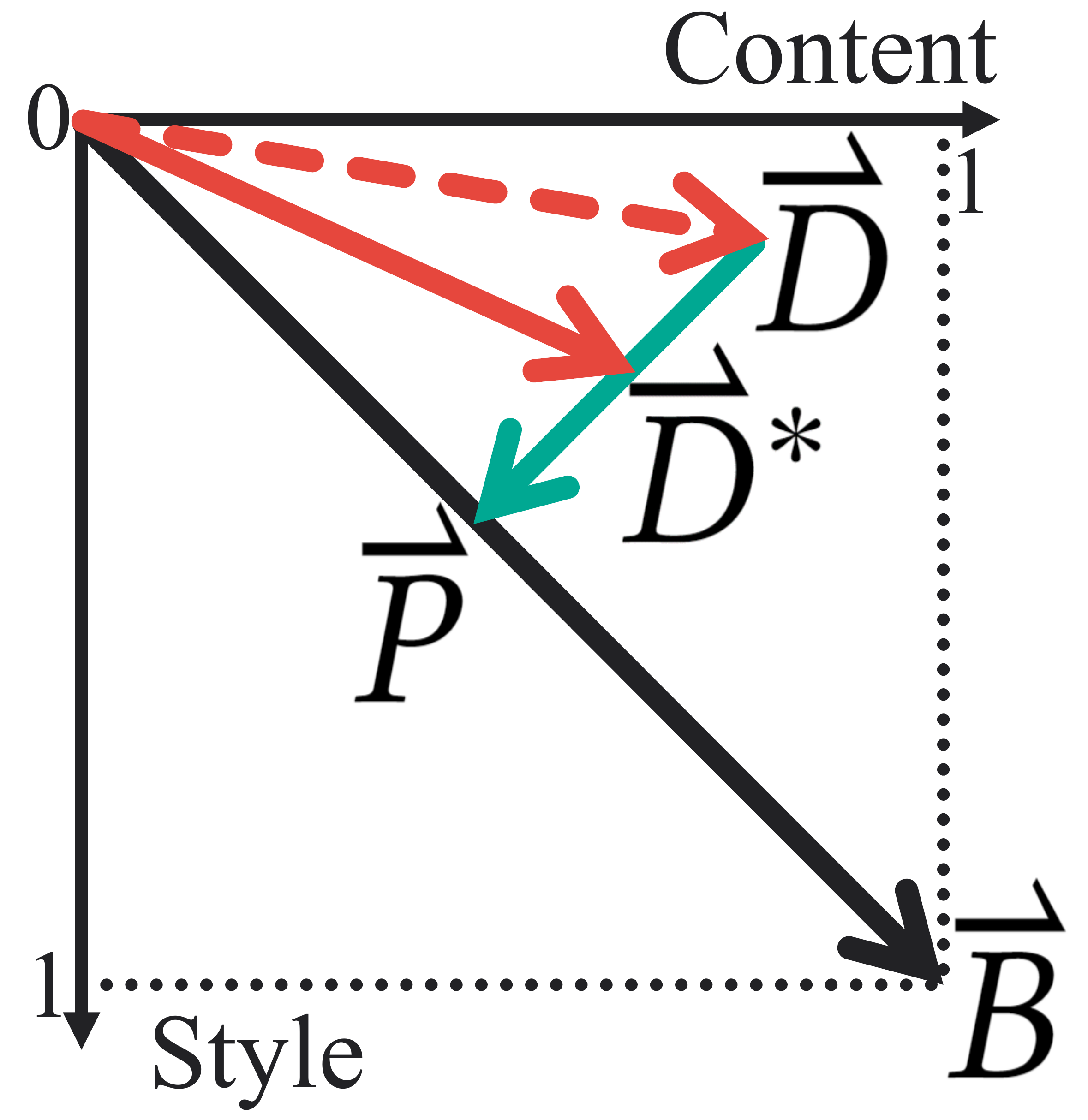} 
    \caption{\label{fig:trunc}Truncation trick.}
\end{wrapfigure}

In areas where the distribution of training data has a low density, it is difficult to learn the true characteristics of the data space and consequently the results may contain artifacts. This remains as an important open problem in learning-based image synthesis and was also observed in our 2D content-style manifold of the latent space. One way to alleviate this situation is to apply a 'truncation trick' \cite{marchesi2017megapixel,pieters2018comparing,kingma2018glow,brock2018large,karras2019style} in order to shrink the latent space and control the trade-off between image diversity and visual quality. We adopted a similar strategy and made some modifications to our framework to ensure reasonable outcomes.

\begin{figure}[h]
    \includegraphics[width=\linewidth]{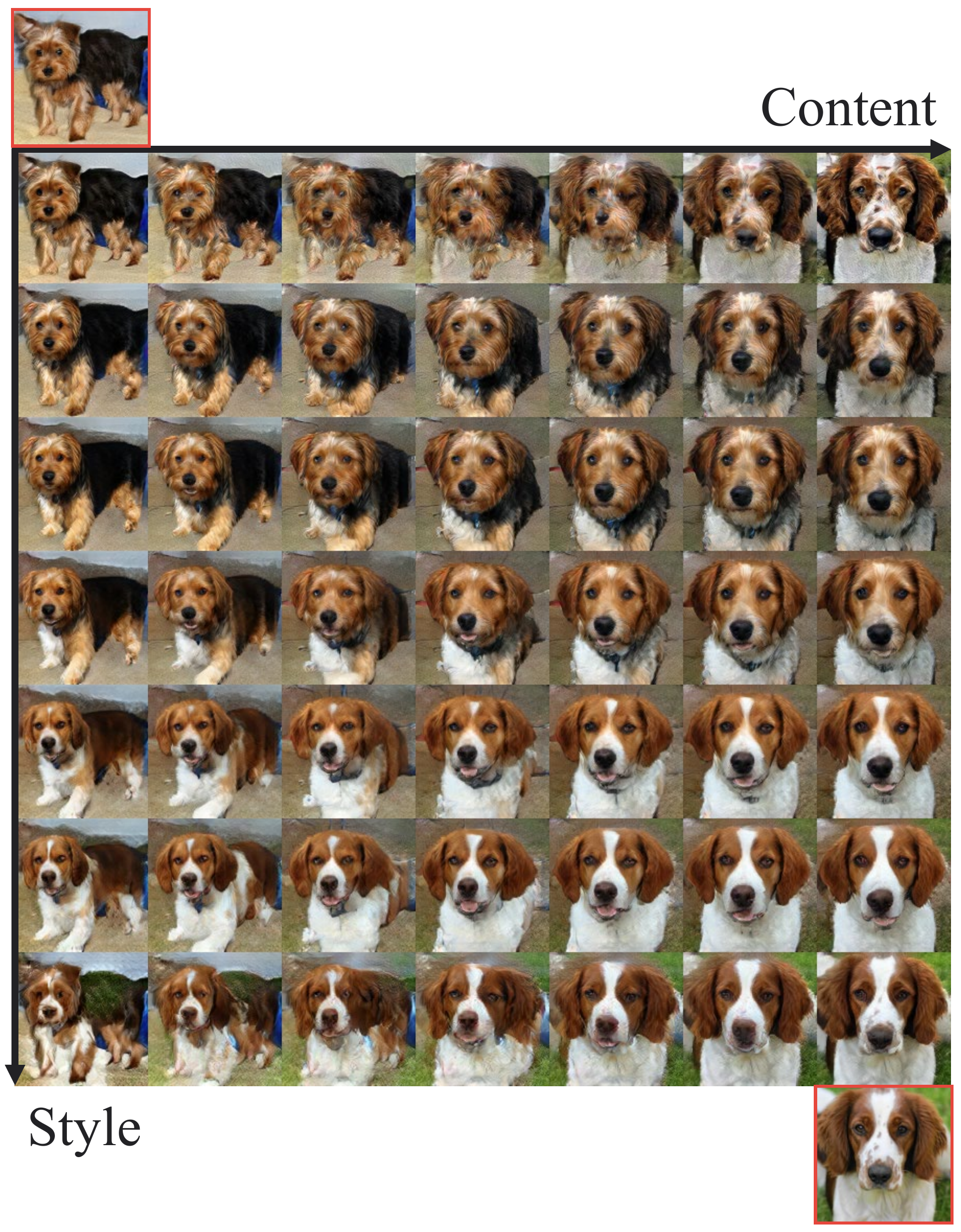} 
    \caption{Visualization of the 2D content-style manifold without applying the truncation trick. Artifacts are observed near the boundary of the manifold.}
    \label{fig:no_trunc}
\end{figure}

The images generated near the basic transition axis tended to follow the distribution assumed by the training data well and consequently had good quality. On the other hand, artifacts are likely to occur near the manifold boundary away from the basic transition axis where content and style components are swapped. Therefore, our truncation strategy is to shrink the 2D content-style manifold toward the diagonal axis where the basic transition is dominant by changing parameter $\tau \{ \tau \in \mathbb{R}: 0\leq\tau\leq 1 \}$. The shrunken manifold is illustrated in Figure \ref{fig:trunc} and can be computed as follows.

\begin{align} \label{eq:truncation}
\begin{split}
\overrightharp{D}^* &=(\alpha_c^*, \alpha_s^*) \\
                    &= \overrightharp{D} + \tau \overrightharp{P} \\
                    &= \overrightharp{D} + \tau \Big( \frac{|\overrightharp{D}|}{|\overrightharp{B}|} \cos{\theta}\overrightharp{B} - \overrightharp{D} \Big) \\
                    &= \overrightharp{D} + \tau \Big( \frac{\overrightharp{D} \cdot \overrightharp{B}}{|\overrightharp{B}|^2} \overrightharp{B} - \overrightharp{D} \Big) \\
                    &= \left( \alpha_c + \tau \Big(\frac{\alpha_s - \alpha_c }{2}\Big), \alpha_s + \tau  \Big(\frac{\alpha_c - \alpha_s }{2}\Big) \right)
\end{split}
\end{align}

where $\overrightharp{D}=(\alpha_c, \alpha_s)$ is an arbitrary vector for a disentangled transition and $\overrightharp{B}=(1, 1)$ is the basic transition vector from input image $x_A$ to $x_B$ on the 2D content-style manifold. $\overrightharp{P}$ is the projection of $\overrightharp{D}$ onto $\overrightharp{B}$. The remapped $\alpha_c^*$ and $\alpha_s^*$ from Equation \ref{eq:truncation} are used to produce a conservatively morphed image for the disentangled transition. Figure \ref{fig:no_trunc} shows the results obtained before applying the truncation trick to the disentangled transition result shown in Figure \ref{fig:disentangled}.

\end{document}